\documentclass[nohyperref]{article}

\usepackage{microtype}
\usepackage{graphicx}
\usepackage{subcaption}
\usepackage{enumitem}
\usepackage{booktabs} 
\usepackage{hyperref}

\usepackage[accepted]{icml2022}

\usepackage{amsmath}
\usepackage{amssymb}
\usepackage{mathtools}
\usepackage{multirow}
\usepackage{amsthm}
\usepackage{wrapfig}
\usepackage[nice]{nicefrac}
\usepackage[normalem]{ulem}
\useunder{\uline}{\ul}{}
\usepackage[capitalize,noabbrev]{cleveref}
\usepackage{tikz}
\newcommand*\circled[1]{\tikz[baseline=(char.base)]{
            \node[shape=circle,draw,inner sep=.5pt] (char) {#1};}}

\DeclareMathOperator*{\argmax}{arg\,max}
\theoremstyle{plain}
\newtheorem{theorem}{Theorem}[section]
\newtheorem{proposition}[theorem]{Proposition}

\theoremstyle{definition}
\newtheorem{definition}[theorem]{Definition}
\newtheorem{assumption}[theorem]{Assumption}
\theoremstyle{remark}

\newtheorem*{claim*}{Claim}

\usepackage[textsize=tiny]{todonotes}

\newcommand\norm[1]{\lVert#1\rVert}

\begin{document}
\icmltitlerunning{Orchestrating Self-Supervised Federated Learning}

\twocolumn[
\icmltitle{Orchestra: Unsupervised Federated Learning via Globally Consistent\\ Clustering}


\icmlsetsymbol{equal}{*}

\begin{icmlauthorlist}
\icmlauthor{Ekdeep Singh Lubana}{umich}
\icmlauthor{Chi Ian Tang}{ucam}
\icmlauthor{Fahim Kawsar}{blabs}
\icmlauthor{Robert P. Dick}{umich}
\icmlauthor{Akhil Mathur}{blabs}
\end{icmlauthorlist}

\icmlaffiliation{umich}{EECS Department, University of Michigan, Ann Arbor, USA}
\icmlaffiliation{ucam}{University of Cambridge, UK}
\icmlaffiliation{blabs}{Nokia Bell Labs, Cambridge, UK}

\icmlcorrespondingauthor{Ekdeep Singh Lubana}{eslubana@umich.edu}

\icmlkeywords{Federated learning, Self-Supervised learning}

\vskip 0.3in
]
\printAffiliationsAndNotice{Work performed while the first two authors were interns at Nokia Bell Labs, UK.} 

\begin{abstract}
Federated learning is generally used in tasks where labels are readily available (e.g., next word prediction). Relaxing this constraint requires design of unsupervised learning techniques that can support desirable properties for federated training: robustness to statistical/systems heterogeneity, scalability with number of participants, and communication efficiency. Prior work on this topic has focused on directly extending centralized self-supervised learning techniques, which are not designed to have the properties listed above. To address this situation, we propose \emph{Orchestra}, a novel unsupervised federated learning technique that exploits the federation's hierarchy to orchestrate a distributed clustering task and enforce a globally consistent partitioning of clients' data into discriminable clusters. We show the algorithmic pipeline in Orchestra guarantees good generalization performance under a linear probe, allowing it to outperform alternative techniques in a broad range of conditions, including variation in heterogeneity, number of clients, participation ratio, and local epochs.
\vspace{-10pt}
\end{abstract}

\section{Introduction}
\label{sec:intro}
\emph{Federated Learning (FL)}~\cite{floriginal} enables collaborative training of machine learning models while avoiding transfer of raw data from clients to server. As remarked by Li et al.~\yrcite{lisurvey}, recent work on this topic focuses on addressing the issues of statistical and systems heterogeneity~\cite{fedprox, scaffold, dirchsplit, noniidfl, flgn}; achieving scalability, privacy, and fairness for participating clients~\cite{largecohort, flsecureagg, fairallocation, ditto, flmultitask}; and improving communication efficiency~\cite{feddyn, flcomms, fedopt, oort}. However, existing FL algorithms generally assume a participating client holds high quality labels that can be used for gradient-based local training~\cite{floriginal}. In \emph{cross-silo} settings, where large organizations collaborate to train a model~\cite{flsurvey}, one can expect each client has an expert to locally label their data. However, in the more constrained scenario of \emph{cross-device} FL~\cite{flsurvey}, clients are generally edge devices and users need to actively interact with the device to label their data locally. This interaction can be hard to arrange beyond specific applications (e.g., next word prediction), thereby hindering FL's adoption with more complex modalities such as vision.  

\begin{figure}
\centering
\includegraphics[width=0.9\columnwidth]{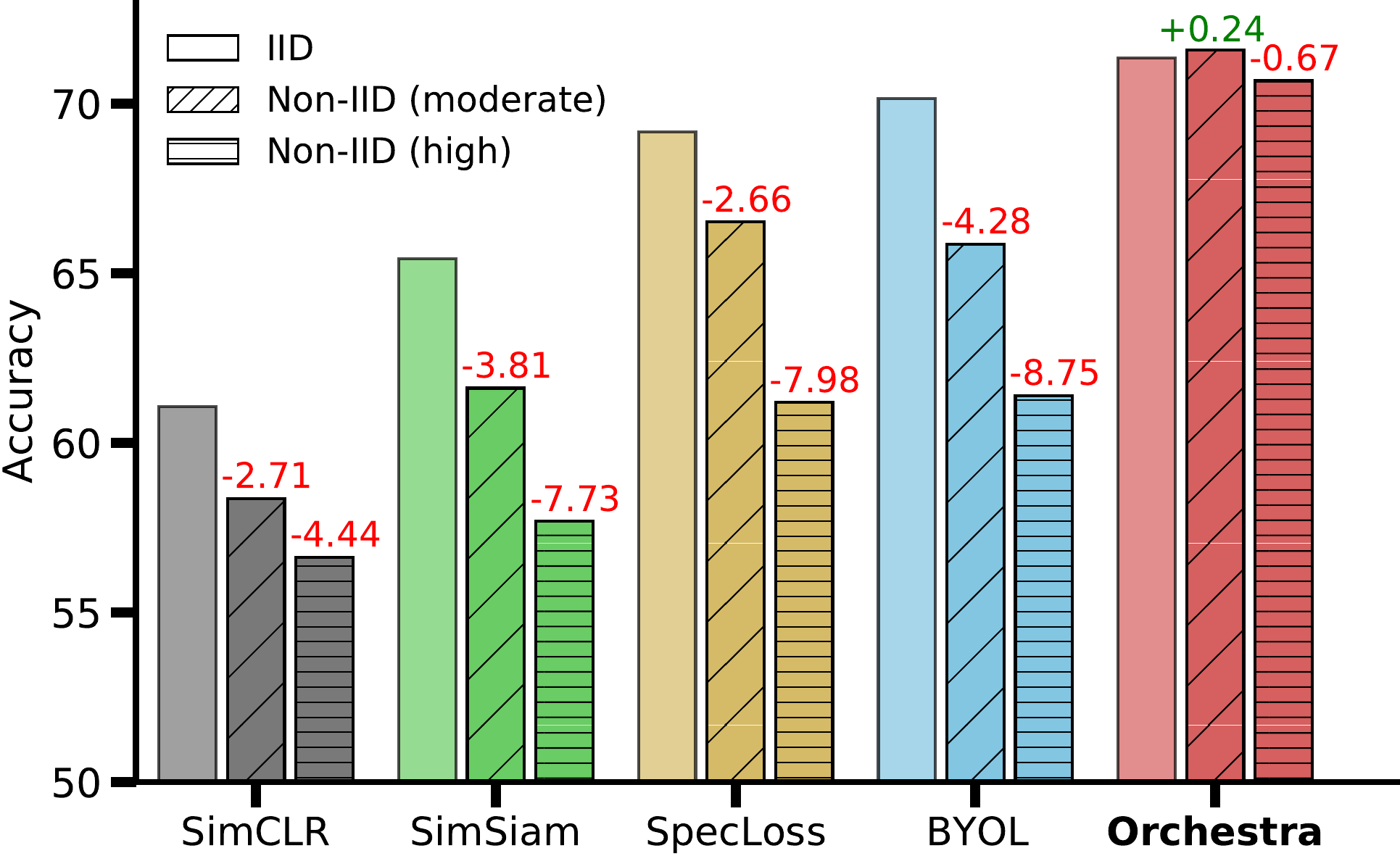}
\vspace{-3mm}
\caption{\textbf{Robustness to Heterogeneity.} We use CIFAR-10 and 100 clients to compare our proposed method (Orchestra) against federated versions of centralized SSL techniques~\cite{simclr, simsiam, specloss, byol}. Under the linear probe protocol~\cite{simclr}, we see that these direct extension methods are sensitive to heterogeneity, while Orchestra remains robust and achieves better absolute accuracy.}
\vspace{-20pt}
\label{fig:intro_impact_of_hetero}
\end{figure}

One possible solution to address this problem is use of unsupervised representation learning algorithms alongside federated training. Some prior works have tried this route by developing direct extensions of self-supervised learning (SSL) techniques from centralized settings, such as SimCLR~\cite{simclr}, BYOL~\cite{byol}, and SimSiam~\cite{simsiam}. However, by using stateful clients~\cite{flcollab, fldivssl}, requiring large batch-sizes~\cite{ssfl}, or sharing representations across clients~\cite{fedmoco, fedca}, these methods are either \emph{only applicable in the cross-silo setting} or \emph{undermine clients' privacy} by enabling inversion of representations~\cite{invertcnn, invertautoregressive} (see also Appendix, Tab.~\ref{table:rw}). If one addresses these limitations by removing state-based operations and constraining training to local data, these methods become mere applications of centralized SSL objectives with federated training. Since centralized SSL techniques are known to be sensitive to heavy-tailed gradient distributions~\cite{dnc} and require large batch-sizes~\cite{simclr}, it is unlikely their direct extensions will function well in the high heterogeneity and resource constrained setting of \emph{cross-device} FL. We demonstrate this behavior in \autoref{fig:intro_impact_of_hetero}, where we show direct extension methods lose noticeable performance with increased heterogeneity in a cross-device FL setting. 

To resolve the limitations noted above and tackle the lack of labelled data in FL, we argue an unsupervised learning framework needs to be developed that is mindful of the challenges seen in federated training. We thus propose \emph{Orchestra}, a novel clustering-based SSL technique that exploits the federation's hierarchy to orchestrate a global partitioning of data distributed across participating clients. Our clustering based perspective arises out of a generalization analysis of models capable of clustering distributed client data into discriminable, low-similarity partitions. As we show, such a model necessarily has a small test error and its performance \emph{generally improves} with \emph{increase} in heterogeneity (e.g., see \autoref{fig:intro_impact_of_hetero}). To exploit this result, we propose to use the server to compute centroids capable of partitioning clients' data into a predefined number of clusters, subsequently asking the participating clients to use these centroids and locally train a model that enforces a sample's cluster assignments on its augmentations. Experiments show that Orchestra scales well with number of clients, achieves strong communication-efficiency, and thrives under heterogeneity. While our focus in this work is resource-constrained, cross-device FL settings, we find Orchestra also outperforms prior works in cross-silo settings. Our primary contributions follow.
\begin{itemize}[noitemsep,topsep=-1pt,leftmargin=6.5pt]
    \item \textbf{Orchestrating Unsupervised FL:} We propose Orchestra, an unsupervised learning technique that addresses the lack of labelled data in FL. The theoretical results motivating our method are discussed in \S\ref{sec:generalization} and its practical implementation is provided in \S\ref{sec:method}.\vspace{2pt}
    \item \textbf{Unsupervised Hyperparameter Tuning:} Since FL algorithms can be sensitive to training hyperparameters~\cite{scaffold, flhparams}, we propose a tuning method for self-supervised FL techniques in \S\ref{sec:tuning}. Our method relies on unlabelled data and finds configurations that yield high (low) representational similarity for related (unrelated) samples.\vspace{2pt}
    \item \textbf{Extensive Empirical Analysis (\S \ref{sec:experiments}):} We extensively compare Orchestra with several federated versions of centralized SSL techniques. We show, unlike Orchestra, direct extension techniques are often sensitive to several important FL parameters (e.g., participation ratio, local epochs). 
\end{itemize}

\section{Preliminaries}
\label{sec:related}

\textbf{Self-Supervised Learning:} SSL is a recent paradigm in unsupervised learning wherein either an application-relevant signal is promoted by predicting properties of the data or an application-irrelevant signal is discarded by discriminating perturbed data~\cite{multiviewssl}. In vision, examples of predictive tasks include colorization~\cite{colorization}, rotation prediction~\cite{rotnet}, or predicting patch permutations~\cite{jigsaw}. Due to high redundancy in visual data, defining task-relevant signal can be difficult and hence predictive tasks rarely yield good results~\cite{contextpred, ylecun}. In contrast, discriminative SSL tasks have revolutionized visual pretraining. Such tasks train the model to enforce invariances to artificial transformations defined using data augmentations, enabling good performance if these transformations encode task-relevant priors~\cite{featurelearning, infomin}. Popular techniques are based on contrastive learning~\cite{simclr, moco, specloss}, similarity-promotion~\cite{byol, simsiam}), redundancy reduction~\cite{btwins}, and clustering~\cite{sela, swav}. 

As mentioned in \S\ref{sec:intro}, several FL papers directly extend the centralized SSL techniques listed above, but their methods are either only applicable in cross-silo settings due to use of stateful clients and large batch sizes, or undermine clients' privacy by sharing representations across clients (see also Table~\ref{table:rw}). We highlight a notable recent exception by Lu et al.~\yrcite{unsupfl}, who assume the server has knowledge of client-level class priors, allowing it to retrieve the exact labels by solving a classic label proportions problem~\cite{labelproportions}. This strong assumption \emph{may} be justifiable in cross-silo settings if participating clients actively agree to share information about expected class priors, but cannot be met in the cross-device setting, where clients are passive.

\textbf{Clustering and Representation Learning:} Prior work has studied the task of clustering a predefined set of vectors distributed across a federated network~\cite{fedcluster} or clustering clients for designing better client selection strategies~\cite{ifca}. In contrast, our work addresses the problem of learning clustering-friendly representations~\cite{towardskmeans}. This problem is known to be difficult even in the centralized setting due to the existence of degenerate solutions, requiring either several expensive reassignment steps~\cite{deepcluster, deepclusterv2}, degeneracy regularization with autoencoders~\cite{depict, deepkmeans}, or partition constraints on large prototype memories~\cite{partitionconfidencemax, onlinedeepclustering}.

For the reasons listed above, we highlight that clustering-based centralized SSL methods like SeLa~\cite{sela} or SwAV~\cite{swav} are not directly applicable to federated settings because they avoid degenerate solutions using periodic cluster re-assignments every few training steps on a large memory module with 4K--1M samples, while also assuming a uniform class prior on the data. Since communication is expensive in FL, each client has only a few samples, and class priors are highly non-uniform. Such requirements make SeLa and SwAV infeasible for federated learning. We note that Orchestra avoids problems noted above by exploiting the federation's hierarchy. Specifically, Orchestra uses the server in a federated manner to find centroids that can partition the clients' data into discriminable clusters. The clients then locally solve an unsupervised clustering problem alongside a predictive SSL task to avoid degenerate solutions. Together, these steps allow Orchestra to learn ``clusterable'' representations.

\textbf{Theory of SSL:} Recently, substantial advances have been made towards demystifying SSL methods from the perspectives of learning theory~\cite{arora}, information theory~\cite{multiviewssl}, causality~\cite{stylecontent}, and dynamical systems~\cite{sslnocontrast}. We use tools proposed in these works to motivate principles behind Orchestra. Specifically, our analysis in \S\ref{sec:generalization} is based on recent works by HaoChen et al.~\yrcite{specloss} and Wei et al.~\yrcite{selftraining}, who derive a holistic framework that allows analysis of generalization error of an unsupervised model. To derive this result, the authors assume there exists a classifier that is able to predict the label of an augmented sample, given the original sample, up to a small error. By being able to predict the labels of both an augmented sample and an original sample (a no-transform augmentation), this assumption implies there exists a latent space where related samples are sufficiently similar and underlying classes are sufficiently dissimilar.

\textbf{Notations}: Before continuing, we define our settings. Assume we have $N > 2$ samples $X \sim \mathcal{X}$ that are distributed across $K$ clients. We assume a ground-truth labelling function $Y(x): \mathcal{X} \to [M]$. The $i^{\text{th}}$ sample $x^{(i)} \in X$ is assigned to client $k^{(i)} \in [K]$; client $k$ has $N^{k}$ samples denoted $X^{k}$. We define a stochastic augmentation function $\mathcal{T}: \mathcal{X} \to \mathcal{\tilde{X}}$ that transforms its input $x$ to the space of augmented samples by randomly selecting a transform from a large, finite set of predefined transformation functions. The set of augmented samples is denoted as $\tilde{X} := \{\mathcal{T}(x): x \in X\}$. We define a parametric representation function $f: \mathcal{X} \to \mathrm{R}^{D}$ and compute its error under a linear probe as $\mathcal{E}(f) := \min_{W} \mathbb{E}_{x\in\mathcal{X}}[\argmax(W^{T}f(x)) = y(x)]$. Note this definition uses the \emph{optimal} linear classifier for a distribution, making its guarantees stronger than a linear probe derived from training data only. The set of representations on a set $\chi$ is denoted as $R_{\chi} = \{f(x): x \in \chi\}$. We use $\mathcal{C}(\mathcal{B}, G)$ to denote a clustering algorithm that returns $G$ clusters with centroids $\mu \in \mathrm{R}^{D \times G}$ on its input $\mathcal{B}$. $\mu_{g}$ denotes centroid of cluster $g$. Cluster assignment probabilities are computed as $P_{f}(x) := \sigma(s_{f}(x, \mu))$, where $\sigma(.)$ denotes the softmax function and $s_{f}(x, \mu) = \mu^{T}f(x)$. $\mathcal{H}(., .)$ denotes cross-entropy between two discrete distributions. $\mathcal{F}$ denotes a hypothesis class that has a global minimizer of the following loss \cite{specloss}: $L_{\text{spec}}=-2\mathbb{E}_{x\in X, \tilde{x}\sim\mathcal{T}(x)}[s_{f}(x, \mu)^{T}s_{f}(\tilde{x}, \mu)]+\mathbb{E}_{x, y\in X}[(s_{f}(x, \mu)^T s_{f}(y, \mu))^{2}]$. 
\section{Clusterability: Symphony Behind Orchestra}
\label{sec:generalization}
In this section, we motivate our reasons for using clustering as the principle behind Orchestra. We begin by analyzing an intentionally \emph{idealized framework} where clients are allowed to share their representations with the server. Our goal is to determine whether evaluating the ``clusterability'' of representations of a federated model trained in an unsupervised manner can provide insights into the quality of the model. To this end, we first define the following.
\begin{definition}\label{defn:mixing} ($\delta$, Inter-Cluster Mixing)
Assume we compute $G$ clusters on a dataset $\mathcal{B}$. Then, inter-cluster mixing is defined as $\delta := \max_{g \in G} \max_{b \in \{\mathcal{B} - g\}}\left(\mu_{g}^{T}b\right)$.
\vspace{-5pt}
\end{definition}
Analogous to the concept of modularity in pairwise clustering~\cite{modularity}, $\delta$ is a similarity measure that can be used for analyzing partition-based clustering algorithms, which explicitly compute centroids and assignments. A smaller $\delta$ denotes better separation between clusters, indicating better ``clusterability'' of the dataset.

\begin{proposition}
\label{prop:distcluster}
Assume $f\in \mathcal{F}$. Compute $\mathcal{G} > 4M+2$ clusters $\mu = \mathcal{C}(\{R_{X^{k}} : k \in [K]\}, \mathcal{G})$ s.t.\ all clusters are equally sized. Then, if $f$ minimizes $\mathcal{L} := \mathbb{E}_{k \in [K]}\left[\mathbb{E}_{x\in X_{k},\tilde{x}\sim\mathcal{T}(x)}\left[\mathcal{H}\left(P_f(x), P_f(\tilde{x})\right)\right]\right]$, we have
\begin{equation}
\footnotesize
\mathcal{E}(f) < \zeta_{\mathcal{X}} + \mathcal{O} \left(2\delta + (G-1)\delta^2\right).
\vspace{-5pt}
\end{equation}
\end{proposition}
Here $\zeta_{\mathcal{X}}$ is a constant that measures the similarity of latent variables of two classes from distribution $\mathcal{X}$, while $\mathcal{O}$ hides constants that primarily depend on the dataset size $N$. Cluster size constraints are employed to ensure that for all classes in the dataset, there is at least one non-trivially sized cluster that contains samples corresponding to that class. If the class-priors were known beforehand, one could enforce size constraints proportional to them, yielding a tighter bound~\cite{selftraining}. However, in unsupervised settings, such priors are unlikely to be known and hence we are forced to use a uniform prior. \emph{Intuitively, Prop.~\ref{prop:distcluster} says that if the representations learned by $f$ are sufficiently diverse to enable computation of $\mathcal{G}$ ``global'' clusters with small inter-cluster mixing $\delta$, then $f$ must have a small linear probe error. Further, the smaller $\delta$ is (more ``clusterable'' representations), the smaller $f$'s error is}.

Overall, Prop.~\ref{prop:distcluster} gives us a target to optimize for while designing our unsupervised FL technique: we must design a method that minimizes $\delta$ and consequently learns good representations. However, the illustrative method above is not yet practical. In particular, requiring that representations of all data be shared with the server can lead to high communication costs over multiple FL rounds. Further, if the server is semi-honest~\cite{flsurvey}, sharing representations can undermine clients' privacy via inversion of representations~\cite{invertcnn, uinvert}. Thus, we need to design a framework that offers a result similar to Prop.~\ref{prop:distcluster} and is yet practical for federated settings. To this end, we propose to perform a \emph{local clustering operation} on our clients. 

\begin{figure}
\centering
\includegraphics[width=0.9\columnwidth]{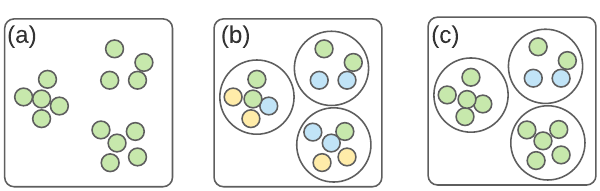}
\vspace{-4mm}
\caption{\textbf{Global Clusters from Different Settings.} Squares denote global clusters. Smaller/larger circles denote samples/local clusters. Colors denote assignments from the idealized setting. (a) Idealized setting. (b) Under low heterogeneity, samples from different ideal clusters exist on a client and can be merged together during local clustering. This leads to global clusters inconsistent with the idealized setting. (c) As heterogeneity increases, samples from fewer idealized clusters exist on each client, thereby reducing inconsistencies and yielding global clusters similar to (a).}
\vspace{-12pt}
\label{fig:localcluster}
\end{figure}

Specifically, we compute $L^{(k)}$ \emph{local} centroids from $N^{(k)}$ representations at each client $k$, share these centroids with the server, and run another clustering operation there to partition the overall set of local centroids into $G$ \emph{global} clusters. This scheme reduces the communication cost per client to just $L^{(k)}$, which can be small if few centroids are used. Further, it provides a general operation where, depending on a system's constraints, different levels of privacy can be added locally without affecting other parts of the pipeline. E.g., for strong privacy guarantees at possibly high loss in utility, locally differentially private (local-DP) clustering methods can be used~\cite{dpbalcan, dpchang}; for slightly weaker guarantees but higher utility, \emph{$K$-anonymous} clustering methods can be used~\cite{kanon1, kanon2, kanon3}. Most importantly, we find local clustering can provide us a generalization guarantee similar to Prop.~\ref{prop:distcluster}.
\begin{proposition}
\label{prop:fedcluster}
Assume $f\in \mathcal{F}$. Denote the set of local centroids as $\mu^{L} = \{\mathcal{C}(R_{X^{k}}, L^{(k)}): k \in [K]\}$ and compute new global centroids $\mu^{G} = \mathcal{C}(\mu^{L}, G)$ s.t.\ all clusters are equally sized. Assume at least a fraction $c$ samples are ``consistently'' assigned, i.e., they match their assignments from the idealized setting. Then, if $f$ minimizes the loss $\mathcal{L} := \mathbb{E}_{k \in [K]}\left[\mathbb{E}_{x\in X_{k},\tilde{x}\sim\mathcal{T}(x)}\left[\mathcal{H}\left(P_f(x), P_f(\tilde{x})\right)\right]\right]$,
\begin{equation}
\footnotesize
\mathcal{E}(f) < \zeta_\mathcal{X} + \mathcal{O} \left(\gamma (1-c^{2}) + (2\delta + (G-1)\delta^2) \right).
\end{equation}
\end{proposition}


Here $\gamma < 1.5$ is a constant and the term $1-c^{2}$ signifies the influence of ``inconsistent'' assignments. In particular, consider a case where samples from multiple idealized clusters are present on a client; during local clustering, such samples can get assigned to the same local cluster. When the centroid of this local cluster is used for global clustering, it inevitably forces samples with different idealized setting assignments to the same global cluster (see \autoref{fig:localcluster}). This increases inconsistencies with the idealized setting (smaller $c$), consequently increasing the upper bound in Prop.~\ref{prop:fedcluster}. 

Interestingly, we observe that heterogeneity in FL setups is beneficial in addressing this challenge! As shown by Dennis et al.~\yrcite{fedcluster}, if the federation has heterogeneity such that a client's data predominantly belongs to only a few clusters (e.g., a smartphone may have several images of the same location), the probability that assignments from centralized clustering match assignments found using global clustering of local centroids approaches 1. If we consider model representations to be a predefined set of vectors distributed across clients, this result directly becomes applicable to our settings: with increase in heterogeneity, $c$ approaches 1 and consequently the bound in Prop.~\ref{prop:fedcluster} matches the bound in Prop.~\ref{prop:distcluster}. \emph{That is, local clustering can use higher heterogeneity to its advantage and achieve guarantees similar to the idealized setting.}

In summary, limitations discovered in the idealized setting (i.e., sharing representations with the server) can be circumvented by relying on local clustering. Furthermore, this general, modular operation exploits heterogeneity to its benefit, preserving guarantees provided by the idealized setting. Thus, our target for optimization via FL remains the same as in Prop.~\ref{prop:distcluster}: we need to develop a training process that produces consistent representations across augmentations, induces sufficient number of global clusters, and has low inter-cluster mixing $\delta$. \emph{Our insight uncovered in Prop.~\ref{prop:fedcluster} is that it is sufficient if these global clusters are computed over local centroids, instead of client representations themselves.}

\begin{figure*}
\centering
\begin{subfigure}{0.53\textwidth}
  \centering
  \centerline{\includegraphics[width=\columnwidth]{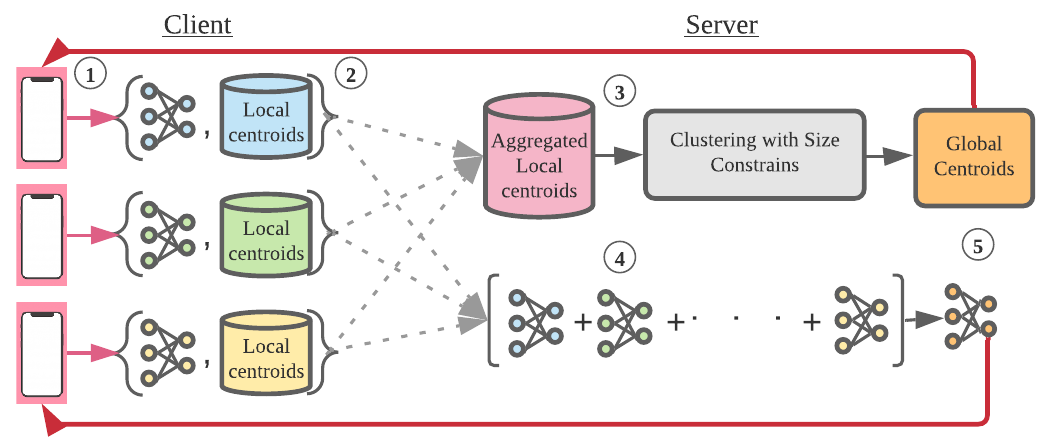}}
  \label{fig:orchestra}
\end{subfigure}%
\begin{subfigure}{0.42\textwidth}
  \centering
  \centerline{\includegraphics[width=\columnwidth]{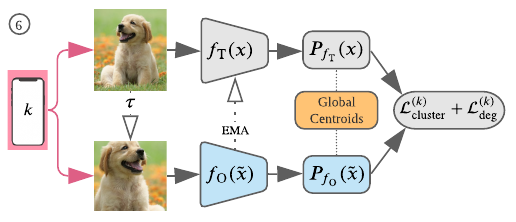}}
  \label{fig:localtrain}
\end{subfigure}
\vspace{-1mm}
\caption{\label{fig:orchestra_all}\textbf{Pipeline:} \circled{1} Orchestra first prompts its clients to compute representations (reps.) on local data. \circled{2} Local centroids are computed via a Sinkhorn-Knopp based clustering algorithm~\cite{otclustering} to constrain clusters to be equally sized. This operation is extremely cheap (0.009\% of client runtime; see \S\ref{subsec:algo}) and enables a K-anonymity privacy guarantee, though local-DP also works well at the expense of some utility (see \S\ref{subsec:privacy}). \circled{3} Local centroids from all clients are aggregated at the server, which again uses Sinkhorn-Knopp clustering~\cite{otclustering} to compute equally-sized global clusters. Note that Sinkhorn-Knopp is used here to satisfy constraints of Prop.~\ref{prop:fedcluster}, not to anonymize the data. \circled{4} Standard FL model averaging step. \circled{5} The global centroids are returned to the clients, who use them for local training. \textbf{Local Training:} \circled{6} An input $x$ is randomly sampled and transformed to $\tilde{x}$. $x$ is converted to rep.\ $f_{\text{T}}(x)$ using an EMA model to enable assignment prediction; $\tilde{x}$ is converted to $f_{\text{O}}(x)$ using the online model. Cluster assignments $P_{f_{\text{T}}}(x), P_{f_{\text{O}}}(\tilde{x})$ are computed by matching reps.\ with global centroids, which are then matched using a cross-entropy loss (\autoref{eq:loss_target}). A predictive SSL task is used to avoid degenerate solutions (\autoref{eq:loss_deg}).}
\vspace{-10pt}
\end{figure*}


\section{Method: How to Conduct an Orchestra}
\label{sec:method}

We now detail the steps underlying our proposed unsupervised FL technique, \emph{Orchestra}. See appendix \S\ref{subsec:algo} for a formal algorithm, implementation details, and github link. 

\textbf{Pipeline:} As shown in Prop.~\ref{prop:fedcluster}, we must ensure $\delta$ is small. To this end, every round, we have clients convert their data into representations and run a clustering algorithm to partition them. The local centroids computed by the clients are then shared with the server, which runs another clustering algorithm on these aggregated centroids. To enforce size constraints from Prop.~\ref{prop:fedcluster} and obtain maximally dissimilar clusters, we use Sinkhorn-Knopp based clustering~\cite{otclustering}. These methods rethink clustering as an optimal transport problem and are generally guaranteed to obtain good approximations of maximally dissimilar clusters. We then communicate the resulting global centroids with the clients, who use them to minimize the cross-entropy between the cluster assignments of a sample and its augmentations. By using the same set of global centroids across all clients, Orchestra's pipeline globally moves cluster members closer to each other, hence reducing $\delta$ every round. More details of the pipeline are provided in \autoref{fig:orchestra_all}. 

\textbf{Local Training:} As noted by prior works in the centralized setting~\cite{depict, deepcluster}, the unstable training dynamics of clustering-based representation learning often yields degenerate solutions. This problem can arise during local training and disallows reduction of $\delta$ beyond a point. We now address this problem to complete the design of Orchestra (see \autoref{fig:orchestra_all}).

1. \textit{Preventing Degenerate Solutions:} During local training, we minimize the KL-divergence between assignments of a sample and its augmentations. Early in the training process, the cluster centroids inevitably correspond to random features and therefore cannot yet yield a sufficiently discriminative signal for training, resulting in degenerate solutions. Centralized methods avoid this problem by adding a degeneracy regularizer in the form of a predictive SSL task that prevents the model from outputting a constant representation. These works primarily use denoising autoencoders~\cite{dec, dac, towardskmeans} for this purpose, but other predictive SSL tasks can also be used, e.g., Rotation Prediction~\cite{rotnet}, Colorization~\cite{colorization}, Jigsaw~\cite{jigsaw}, or Context Prediction~\cite{contextpred}. However, except for rotation prediction, predictive SSL tasks can have high resource requirements because they require larger models (autoencoders, colorization) or computation of representations from several patches of an input (Jigsaw, Context Prediction). Rotation Prediction is best suited to our federated settings because it adds only an extra forward/backward pass and a linear layer worth of memory cost. Thus, every training step, we rotate each sample $x$ in a batch to $\mathcal{R}(x, \theta)$ by sampling an angle $\theta$ from the set $\alpha= \{0^{\circ}, 90^{\circ}, 180^{\circ}, 270^{\circ}\}$ via uniform distribution $\mathcal{U}_4$. A linear layer $W_r \in \mathbb{R}^{D \times 4}$ is then trained to predict the sample's rotation from its representation. Specifically, if $W_{r, i}$ and $\alpha_{i}$ index $W_{r}$ and $\alpha$, we get the following objective.
\begin{equation}
\label{eq:loss_deg}
\footnotesize
    \mathcal{L}_{\text{deg}}^{(k)} = \mathbb{E}_{x \in X^{k}, i \in \mathcal{U}_{4} }\left[\arg\max\left(W_{r,i}^{T}f(\mathcal{R}(x, \alpha_i))\right) = i\right]
\end{equation}

2. \textit{Predicting Assignments:} After training for a few iterations, the model representations change; possibly changing a sample's cluster assignment. This pushes the model to learn varying assignments for the same sample over time, forcing it to predict uniform assignments for all samples~\cite{onlinedeepclustering}. Prior works in centralized settings have proposed to solve this problem by calculating assignments alongside clusters and keeping them fixed until the next clustering operation, which happens every few iterations. In our setting, this solution would require more frequent communication between the server and the clients, which will be expensive. Instead, we propose to use two models: an online model that is trained using gradient descent and another model whose parameters are an exponential moving average (EMA) of the online model: $\text{T}^{t} = m \text{T}^{t-1} + (1 - m) \text{O}$, where $m$ is a scalar close to $1$, $\text{T}$ denotes parameters of EMA model, and $\text{O}$ denotes parameters of the online model. While centralized SSL works also use such EMA models~\cite{byol,dino}, our primary motivation for this design choice is that its representations evolve slowly over time and hence its assignments remain consistent with the original ones. This yields the following loss function for promoting clusterability of representations:
\begin{equation}
\label{eq:loss_target}
\footnotesize
    \mathcal{L}_{\text{cluster}}^{(k)} = \mathbb{E}_{x \in X^{k},\tilde{x}\sim\mathcal{T}(x)}\left[\mathcal{H}\left(P_{f_\text{T}}(x), P_{{f_\text{O}}}(\tilde{x})\right)\right].
\end{equation}
Notice that asymptotically, the target model and online model will share the exact same parameters. Our results in Prop.~\ref{prop:distcluster} and \ref{prop:fedcluster} are primarily designed for this regime, hence they continue to hold for Eq.~\ref{eq:loss_target}. 
\section{Hyperparameters: Tuning our Instruments}
\label{sec:tuning}

Properly tuning hyperparmeters is of paramount importance in FL~\cite{flhparams}. However, due to lack of labelled data, this can be particularly hard for unsupervised FL and can lead to conflicting results. E.g., we note that while Zhuang et al.~\yrcite{flcollab} find BYOL outperforms its competitors in federated settings, He et al.~\yrcite{ssfl} claim it collapses to degenerate solutions. This discrepancy likely stems from the use of a different EMA in the two works (0.99 vs.\ 0.9).

To avoid such inconsistencies and ensure fair comparisons, we propose to tune the hyperparameters of all methods in an unsupervised manner. Specifically, we compute the following two similarity scores:
\begin{equation}
\label{eq:algn_unif}
\footnotesize
\begin{split}
    \text{Align}(f) &= \mathbb{E}_{k \in [K]}\left[\mathbb{E}_{x\in X_{k},\tilde{x}\sim\mathcal{T}(x)}\left[s(x, \tilde{x})\right]\right],\, \text{and} \\
    \text{Unif}(f, \tau) &= -\mathbb{E}_{k \in [K]}\left[\mathbb{E}_{x\in X_{k}}\left[\log \mathbb{E}_{y\in X_{k}}\left[e^{\nicefrac{s(x, y)}{\tau}}\right]\right]\right],\\
\end{split}
\end{equation}
where $s(x, y) = \nicefrac{f(x)^{T}{f}(y)}{\norm{f(x)}\norm{f(y)}}$ denotes cosine similarity of representations and $\tau$ is a vMF distribution parameter~\cite{vmf}. Proposed by Wang and Isola~\yrcite{uniformity}, the two scores are respectively large if the representations are similar for augmentations of a sample (high alignment) and dissimilar across samples (high uniformity). 

\begin{figure}
\centering
\includegraphics[width=\columnwidth]{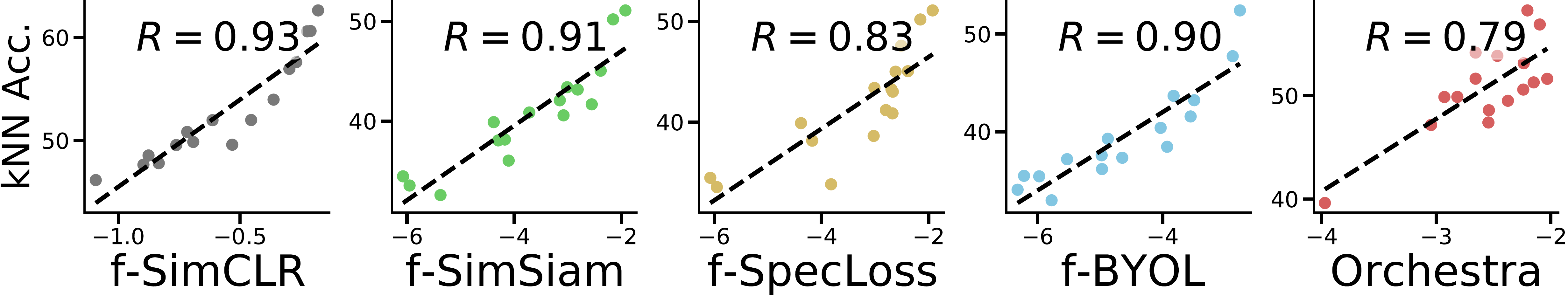}
\vspace{-4mm}
\caption{\textbf{Hyperparameter tuning.} We find linear combination of similarity scores (\autoref{eq:algn_unif}) is highly predictive of kNN accuracy across all methods. Here, x-axis denotes Align + 0.2 * Unif and $\tau$ is set to 0.2, based on Wang and Isola~\yrcite{uniformity}. For all methods, ResNet-18 models are trained on CIFAR-10 using different settings of learning rates (0.3, 0.01, 0.003, 0.001), heterogeneity (10 or $\sim$3.5 classes per client), and number of clients (10 or 100).}
\vspace{-5pt}
\label{fig:tuning}
\end{figure}

Different versions of these scores show up in generalization bounds of SSL methods, under the assumption of task-relevant augmentations~\cite{saunshi_icml, arora, selftraining, sslgen, specloss}. Given this intricate relationship, we argue these scores are likely to be predictive of the quality of model representations. We verify this claim by plotting the kNN accuracy ($k=200$) achieved by different methods in different settings as a function of a linear combination of the two scores. As shown in \autoref{fig:tuning}, across all methods, the similarity scores are highly predictive of achieved accuracy (R = 0.87, on average). Thus, for all methods in our experiments, we propose to choose hyperparameters that maximize the linear combination of the similarity scores above.
\section{Experiments: The Concert}
\label{sec:experiments}
This section compares Orchestra with federated versions of several discriminative SSL methods: SimCLR~\cite{simclr}, SpecLoss~\cite{specloss}, SimSiam~\cite{simsiam}, and BYOL~\cite{byol}. Results on rotation prediction (RotPred)~\cite{rotnet} and supervised FedAvg~\cite{floriginal} are also provided. For cross-silo experiments, we use implementation tricks by recent self-supervised FL~\cite{flcollab} papers to make our baselines even more competitive.

\textbf{Setup.} We use CIFAR-10/-100 datasets. To partition data across $K$ clients, we sample class priors from a Dirichlet distribution~\cite{dirchsplit}. A smaller Dirichlet parameter $\alpha$ yields more heterogeneous splits. All methods are implemented using PyTorch and the Flower framework~\cite{beutel2020flower}. We use ResNet-18 as the backbone architecture; Projector/Predictor architectures follow original papers. Orchestra uses a 2-layer Projector, but no Predictor. We compute 64/128 global clusters and 8/16 local clusters for CIFAR-10/-100. All results are averaged over 3 seeds. Standard deviations are shown in figures, but omitted from tables and deferred to the appendix due to space constraints. Learning rate is tuned for all SSL methods as per \S\ref{sec:tuning}; for FedAvg, we borrow values from Charles et al.~\yrcite{largecohort}. We tune the EMA value $m$ for BYOL. Batch-size is set to 16 (256) for cross-device (cross-silo) settings. Unless stated otherwise, we set $\alpha$ to 0.1, number of local epochs $E$ to 10, communication rounds $C$ to 100, and participation ratio $R$ to 0.5 (1.0) for cross-device (cross-silo) experiments. Please refer to \S\ref{appendix:experiment} for more details on the experiment setup.  

\textbf{Evaluation Protocol.} We primarily use the standard linear probe protocol, where the model is frozen and a linear classifier is learned on top of the backbone~\cite{simclr}. When comparing communication efficiency, we use a kNN accuracy probe~\cite{simsiam}. For semi-supervised evaluation, we fine-tune the entire model using limited labelled data (1\% or 10\% labels).

\subsection{Linear and Semi-Supervised Evaluation}
\label{sec:linear_eval}
\begin{table*}[ht]
\caption{\label{tab:eval_results} Accuracy (\%) in non-IID ($\alpha$=0.1) cross-device (100 clients) and cross-silo (10 clients) settings on CIFAR datasets. For cross-device settings, due to a lack of baselines, we use FL-extensions of several centralized techniques; for cross-silo settings, we follow implementations of these techniques proposed by recent works that use stateful clients and divergence-aware predictor updates~\cite{flcollab}. We evaluate models using the popular linear probe technique and semi-supervised fine-tuning with 1\%/10\% labelled data. Consistent with prior work~\cite{flcollab}, we note that linear probe can outperform semi-supervised evaluation on CIFAR-10.}
\centering
\begin{tabular}{@{}c|cccccc|cccccc@{}}
\toprule
Dataset      & \multicolumn{6}{c|}{CIFAR-10}                                                                                                                                                                                & \multicolumn{6}{c}{CIFAR-100}                                                                                                                                                                                \\ \midrule
Setting      & \multicolumn{3}{c|}{Cross-Device (K = 100)}                                                                               & \multicolumn{3}{c|}{Cross-Silo (K=10)}                                                            & \multicolumn{3}{c|}{Cross-Device (K = 100)}                                                                               & \multicolumn{3}{c}{Cross-Silo (K=10)}                                                             \\ \midrule
	         & \multicolumn{1}{c|}{Linear}         & \multicolumn{1}{c|}{1\%}            & \multicolumn{1}{c|}{10\%}           & \multicolumn{1}{c|}{Linear}         & \multicolumn{1}{c|}{1\%}            & 10\%           & \multicolumn{1}{c|}{Linear}         & \multicolumn{1}{c|}{1\%}            & \multicolumn{1}{c|}{10\%}           & \multicolumn{1}{c|}{Linear}         & \multicolumn{1}{c|}{1\%}            & 10\%           \\ \midrule
f-SimCLR       & \multicolumn{1}{c|}{58.36}          & \multicolumn{1}{c|}{41.95}          & \multicolumn{1}{c|}{44.64}          & \multicolumn{1}{c|}{69.29}          & \multicolumn{1}{c|}{57.76}          & 68.27          & \multicolumn{1}{c|}{34.52}          & \multicolumn{1}{c|}{45.47}          & \multicolumn{1}{c|}{51.88}          & \multicolumn{1}{c|}{44.33}          & \multicolumn{1}{c|}{57.61}          & 67.84          \\
f-SimSiam      & \multicolumn{1}{c|}{61.61}          & \multicolumn{1}{c|}{49.99}          & \multicolumn{1}{c|}{56.43}          & \multicolumn{1}{c|}{75.12}          & \multicolumn{1}{c|}{64.04}          & 72.25          & \multicolumn{1}{c|}{34.96}          & \multicolumn{1}{c|}{47.17}          & \multicolumn{1}{c|}{55.13}          & \multicolumn{1}{c|}{43.16}          & \multicolumn{1}{c|}{53.38}          & 63.19          \\ 
f-SpecLoss     & \multicolumn{1}{c|}{66.51}          & \multicolumn{1}{c|}{55.66}          & \multicolumn{1}{c|}{62.09}          & \multicolumn{1}{c|}{{\ul 80.71}}    & \multicolumn{1}{c|}{70.88}          & {\ul 77.96}    & \multicolumn{1}{c|}{37.60}          & \multicolumn{1}{c|}{47.11}          & \multicolumn{1}{c|}{50.91}          & \multicolumn{1}{c|}{\textbf{56.59}} & \multicolumn{1}{c|}{{\ul 62.15}}    & {\ul 72.09}    \\
f-BYOL         & \multicolumn{1}{c|}{{\ul 65.85}}    & \multicolumn{1}{c|}{{\ul 56.05}}    & \multicolumn{1}{c|}{{\ul 64.15}}    & \multicolumn{1}{c|}{76.08}          & \multicolumn{1}{c|}{{\ul 65.55}}    & 73.18          & \multicolumn{1}{c|}{38.47}          & \multicolumn{1}{c|}{{\ul 52.89}}    & \multicolumn{1}{c|}{{\ul 58.56}}    & \multicolumn{1}{c|}{49.64}          & \multicolumn{1}{c|}{57.34}          & 66.13          \\
Orchestra    & \multicolumn{1}{c|}{\textbf{71.58}} & \multicolumn{1}{c|}{\textbf{60.33}} & \multicolumn{1}{c|}{\textbf{66.20}} & \multicolumn{1}{c|}{\textbf{82.14}} & \multicolumn{1}{c|}{\textbf{71.30}} & \textbf{79.51} & \multicolumn{1}{c|}{\textbf{40.37}} & \multicolumn{1}{c|}{\textbf{54.01}} & \multicolumn{1}{c|}{\textbf{59.07}} & \multicolumn{1}{c|}{{\ul 55.89}}          & \multicolumn{1}{c|}{\textbf{63.73}} & \textbf{73.06} \\\midrule
RotPred (pred) & \multicolumn{1}{c|}{44.44}          & \multicolumn{1}{c|}{34.71}          & \multicolumn{1}{c|}{46.15}          & \multicolumn{1}{c|}{55.68}          & \multicolumn{1}{c|}{45.84}          & 51.32          & \multicolumn{1}{c|}{16.85}          & \multicolumn{1}{c|}{15.79}          & \multicolumn{1}{c|}{19.52}          & \multicolumn{1}{c|}{25.0}           & \multicolumn{1}{c|}{27.64}          & 28.81          \\
FedAvg (sup) & \multicolumn{1}{c|}{80.85}          & \multicolumn{1}{c|}{82.76}          & \multicolumn{1}{c|}{80.34}          & \multicolumn{1}{c|}{79.22}          & \multicolumn{1}{c|}{86.81}          & 87.12          & \multicolumn{1}{c|}{58.71}          & \multicolumn{1}{c|}{59.07}          & \multicolumn{1}{c|}{64.01}          & \multicolumn{1}{c|}{65.59}          & \multicolumn{1}{c|}{62.48}          & 71.59          \\\bottomrule
\end{tabular}
\end{table*}
We first assess the accuracies of different methods using both a linear probe protocol and semi-supervised evaluation. Though our primary motivation in designing Orchestra is cross-device FL, for this experiment, we evaluate its performance in cross-silo settings as well. Our cross-device setting has 100 clients, while the cross-silo setting has 10 clients. $\alpha$ is set to 0.1. Results are shown in \autoref{tab:eval_results}. 

We make several observations. (i) Orchestra often outperforms alternative techniques by a large margin under the linear probe protocol, where primarily the quality of learned representations is evaluated. (ii) Under semi-supervised settings, the gap reduces, but remains high when only 1\% labels are used. (iii) Even though Orchestra was designed with cross-device settings in mind, we find that it also outperforms its competitors in cross-silo settings. We expect Orchestra's performance can be improved further with stateful operations and other enhancements allowed in cross-silo settings. This is left to future work.

\subsection{Attributes of Federated Learning}
\label{sec:fl_experiments}

\begin{figure}
\centering
\begin{subfigure}{0.48\columnwidth}
  \centering
  \centerline{\includegraphics[width=\columnwidth]{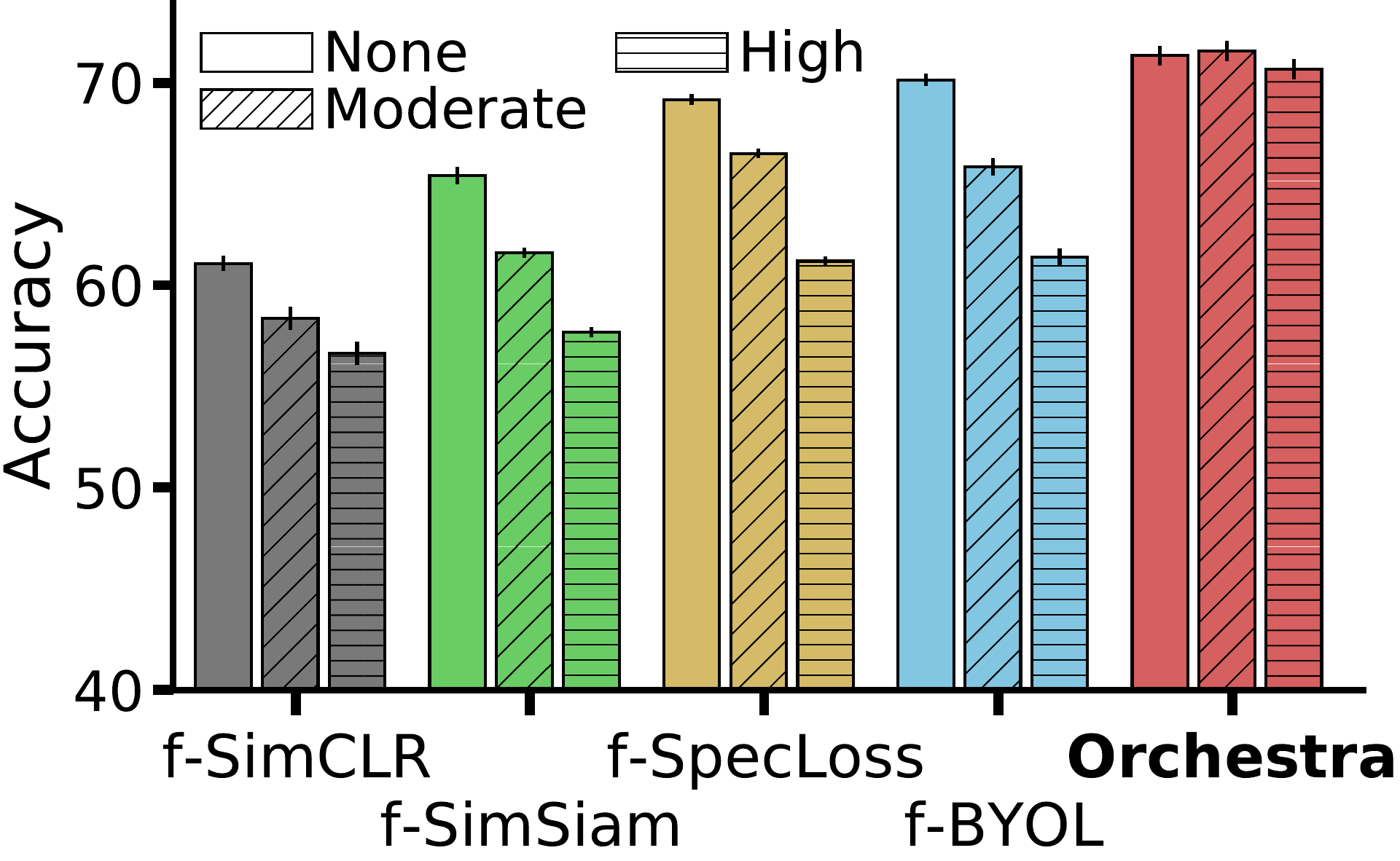}}
\end{subfigure}%
\begin{subfigure}{0.48\columnwidth}
  \centering
  \centerline{\includegraphics[width=\columnwidth]{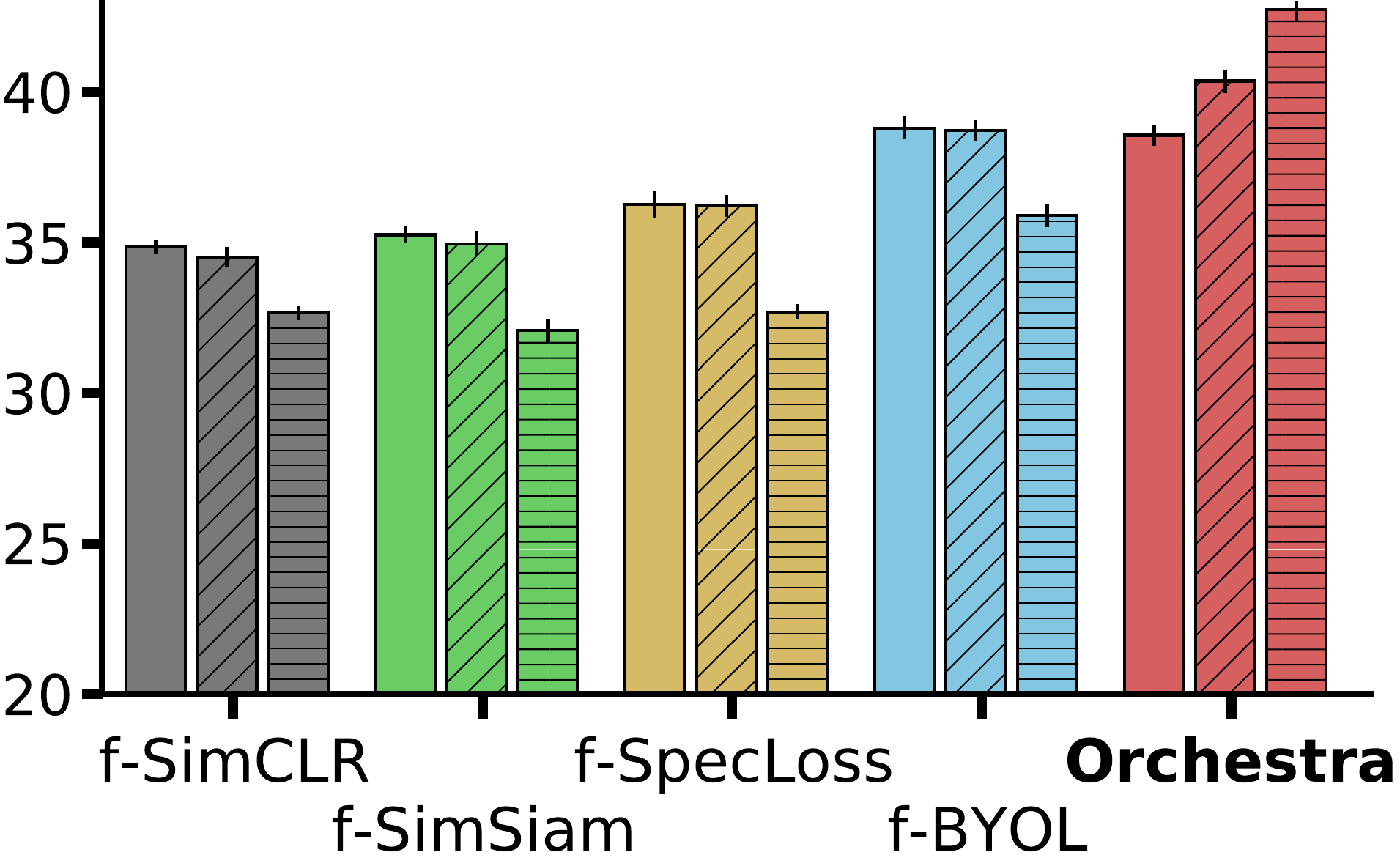}}
\end{subfigure}
\vspace{-2mm}
\caption{\label{fig:exptt_hetero} \textbf{Sensitivity to Statistical Heterogeneity} on CIFAR-10 (left) and CIFAR-100 (right). Except for Orchestra, we find all methods lose performance with increased heterogeneity.}
\vspace{-10pt}
\end{figure}
\begin{figure}
\centering
\begin{subfigure}{0.48\columnwidth}
  \centering
  \centerline{\includegraphics[width=\columnwidth]{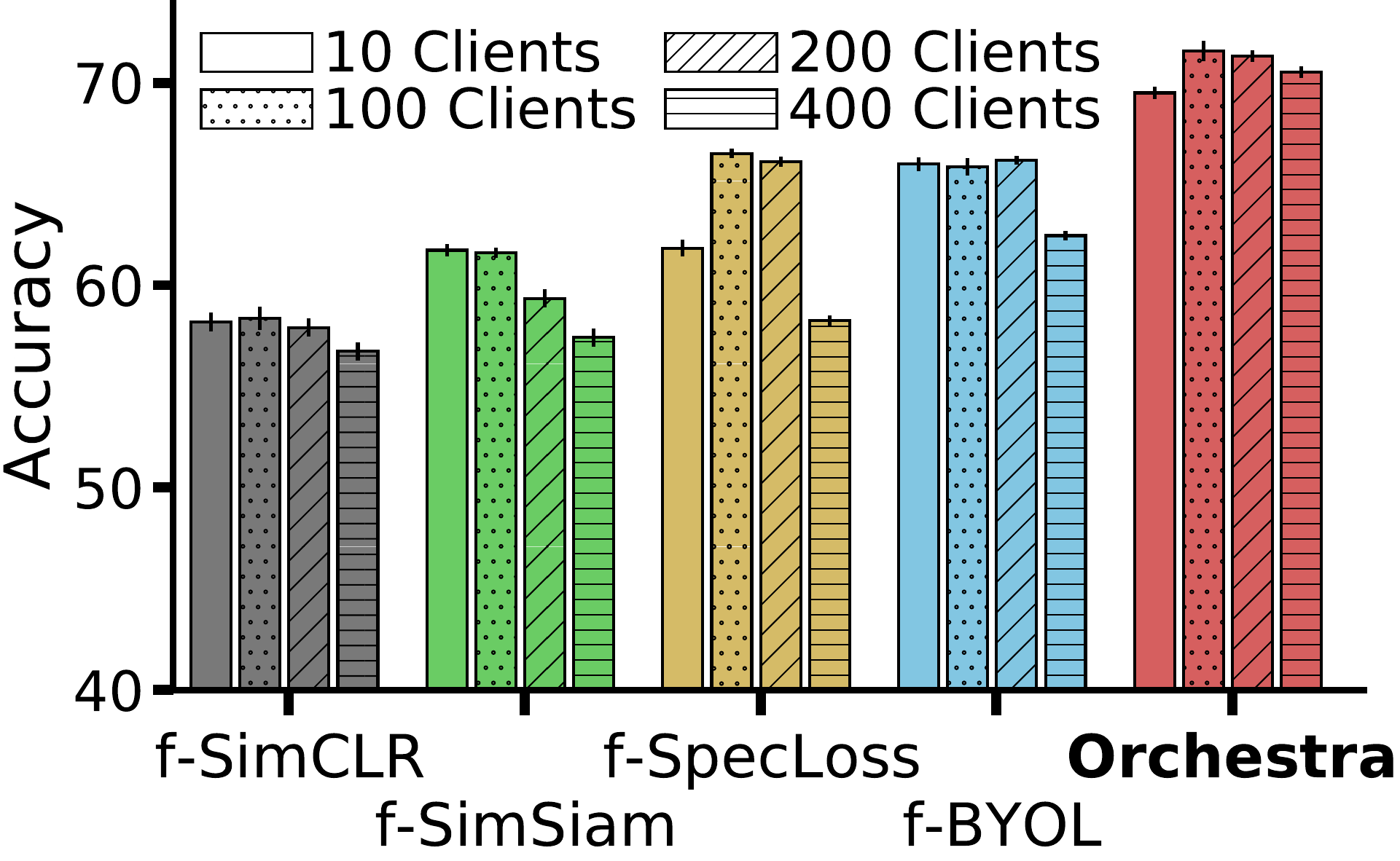}}
\end{subfigure}%
\begin{subfigure}{0.48\columnwidth}
  \centering
  \centerline{\includegraphics[width=\columnwidth]{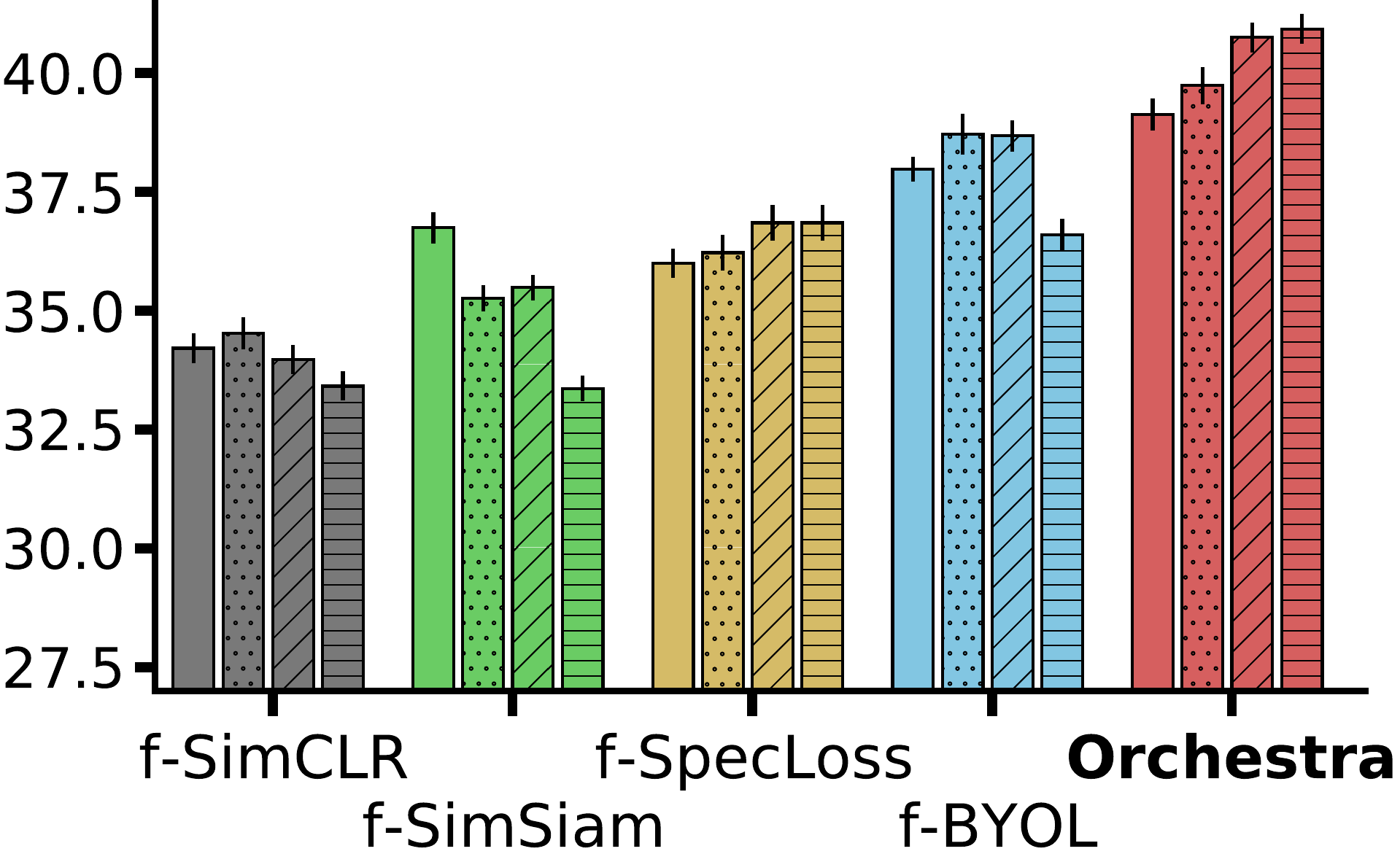}}
\end{subfigure}
\vspace{-2mm}
\caption{\label{fig:exptt_scale} \textbf{Scalability with number of clients} on CIFAR-10 (left) and CIFAR-100 (right). Orchestra outperforms other methods and is generally robust to number of clients in the federation.}
\vspace{-15pt}
\end{figure}
\textbf{Statistical Heterogeneity.} We use 100 clients and analyze three levels of heterogeneity by setting $\alpha$ to 10$^{\text{5}}$ (none/IID), 10$^{\text{-1}}$ (moderate), and 10$^{\text{-3}}$ (high). Results are provided in \autoref{fig:exptt_hetero}. We see that FL-extensions of centralized SSL methods are often sensitive to heterogeneity, especially on CIFAR-10. In contrast, Orchestra is robust and its performance generally improves with increase in heterogeneity. This observation matches our expectation from Prop.~\ref{prop:fedcluster}, where we showed Orchestra's theoretical guarantees improve under increased heterogeneity. We also note that since CIFAR-100 has fewer samples per class, increase in heterogeneity appears to enable greater gains via reduction in inconsistent assignments.

\begin{figure}
\centering
\begin{subfigure}{0.48\columnwidth}
  \centering
  \centerline{\includegraphics[width=\columnwidth]{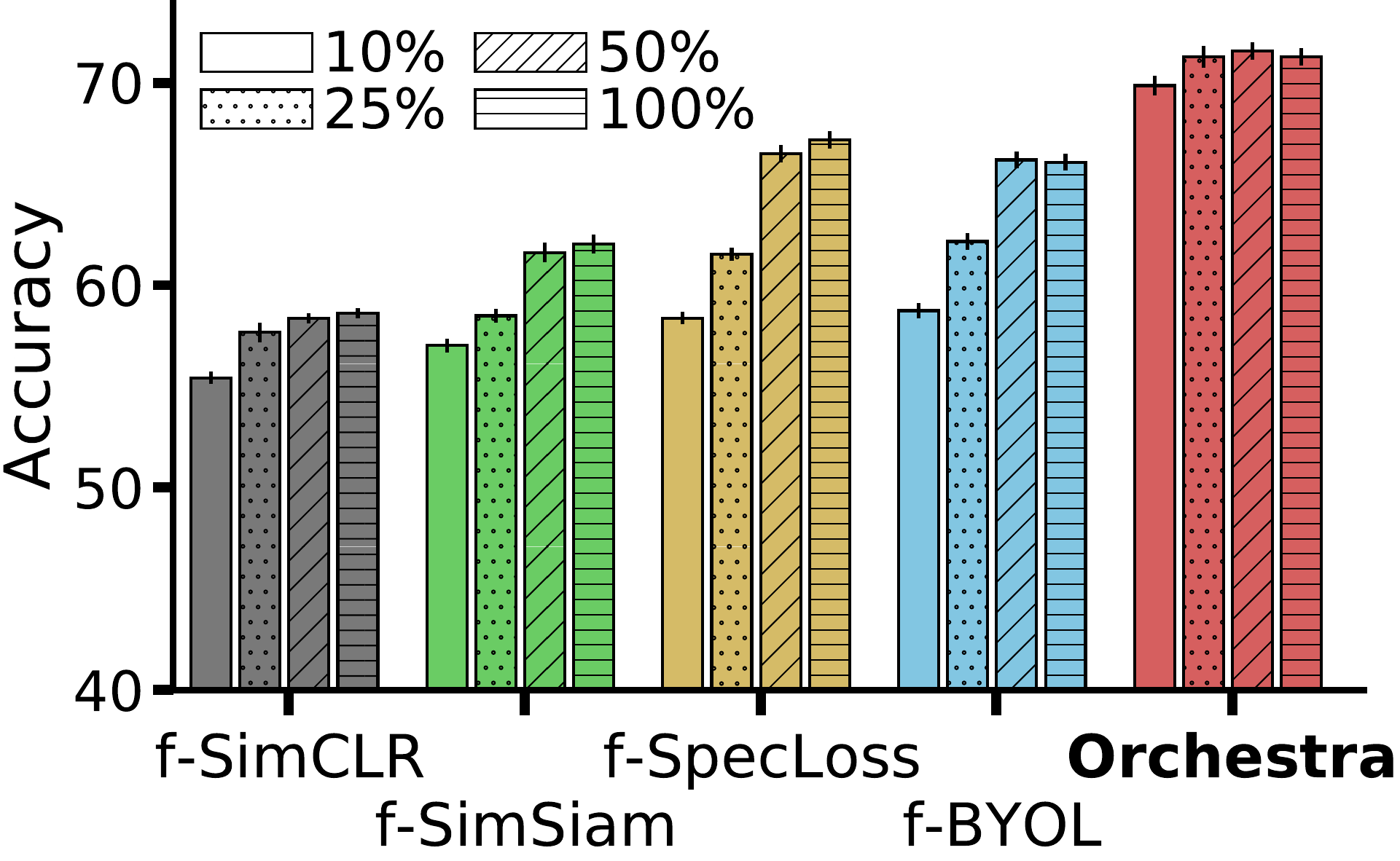}}
\end{subfigure}%
\begin{subfigure}{0.48\columnwidth}
  \centering
  \centerline{\includegraphics[width=\columnwidth]{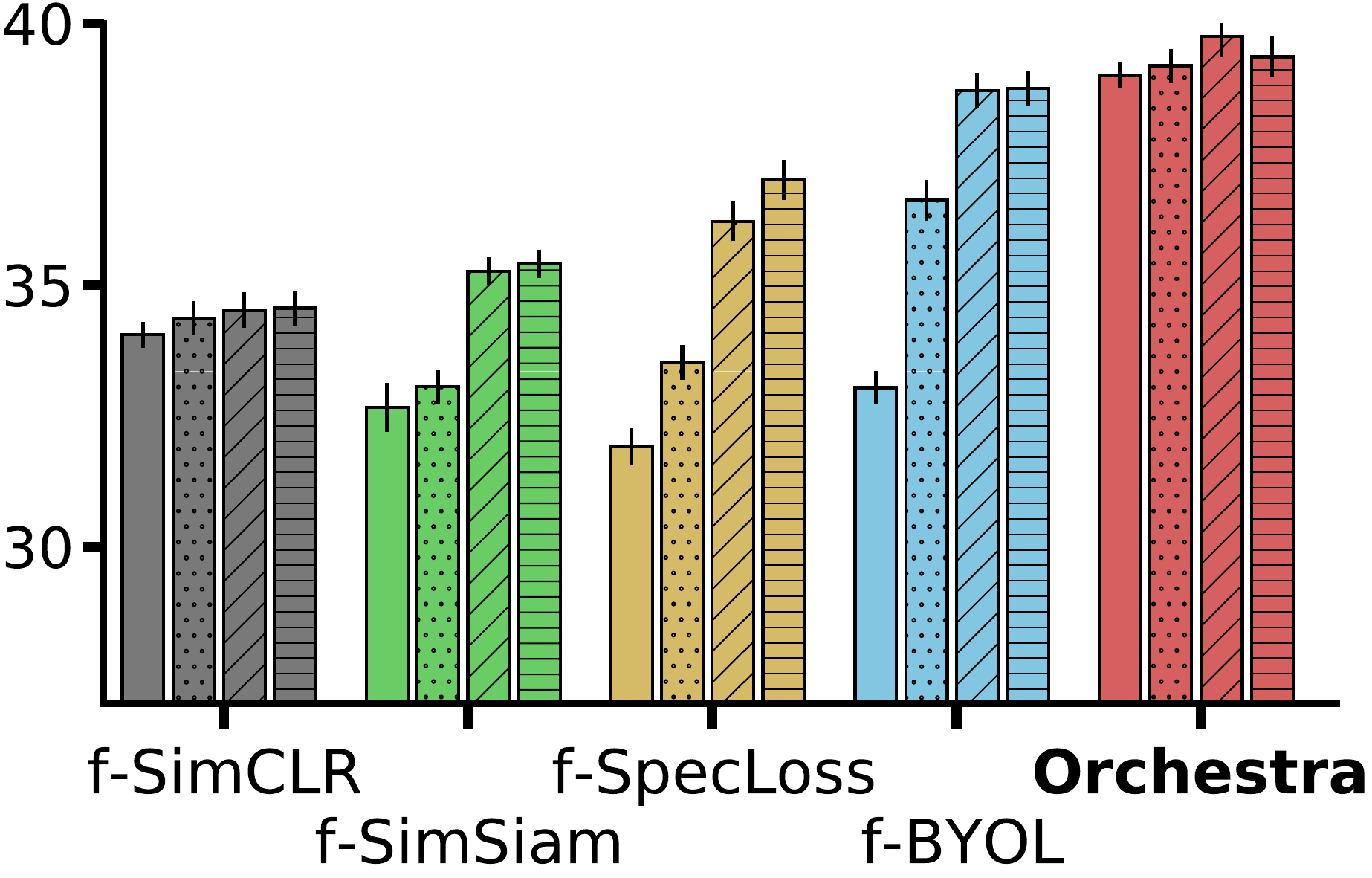}}
\end{subfigure}
\vspace{-2mm}
\caption{\label{fig:exptt_participation} \textbf{Sensitivity to participation ratio} on CIFAR-10 (left) and CIFAR-100 (right). While the accuracies of alternative techniques decrease at smaller participation ratios, we find Orchestra suffers minimal degradation.}
\vspace{-10pt}
\end{figure}
\begin{figure}
\centering
\begin{subfigure}{0.48\columnwidth}
  \centering
  \centerline{\includegraphics[width=\columnwidth]{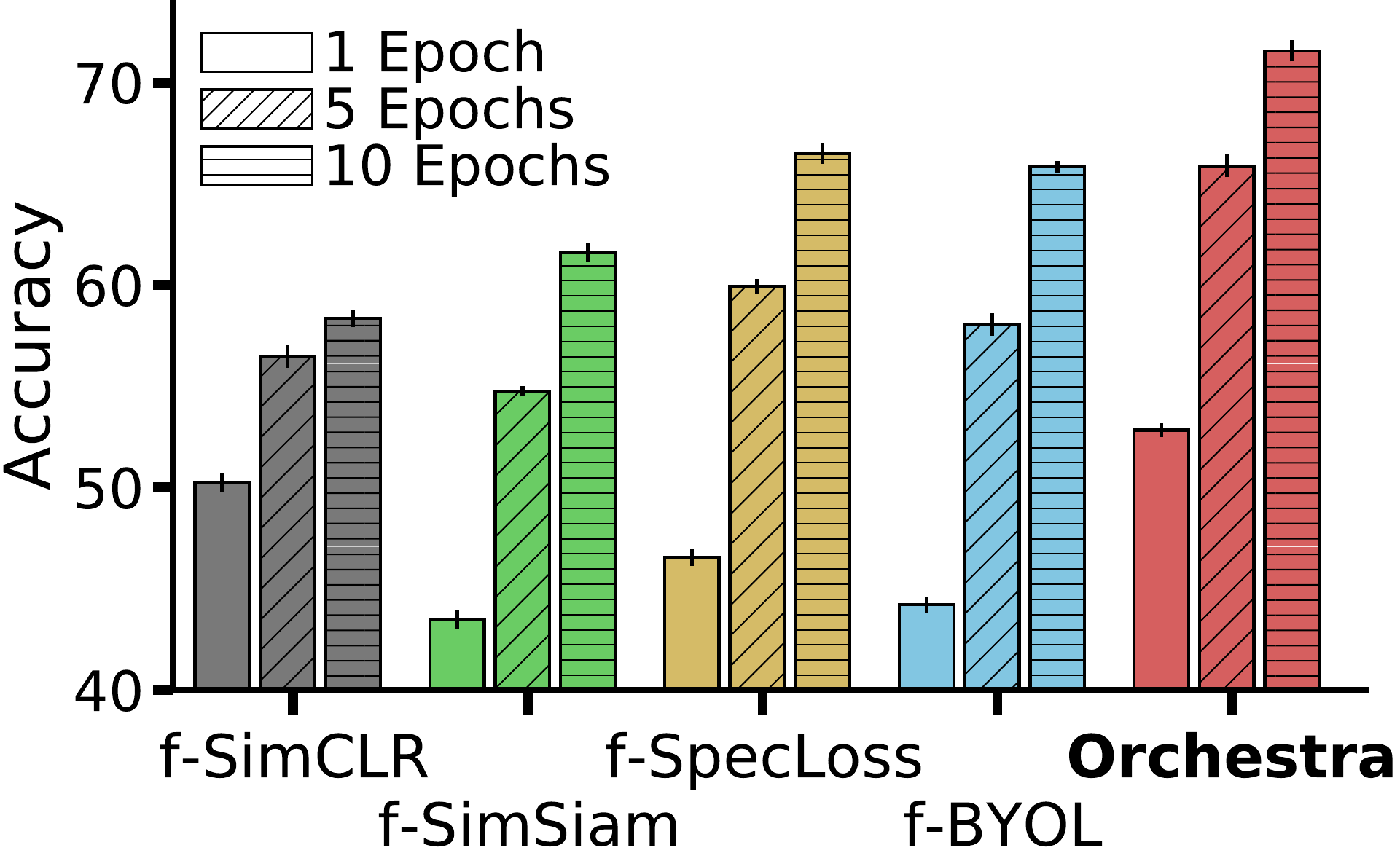}}
\end{subfigure}%
\begin{subfigure}{0.48\columnwidth}
  \centering
  \centerline{\includegraphics[width=\columnwidth]{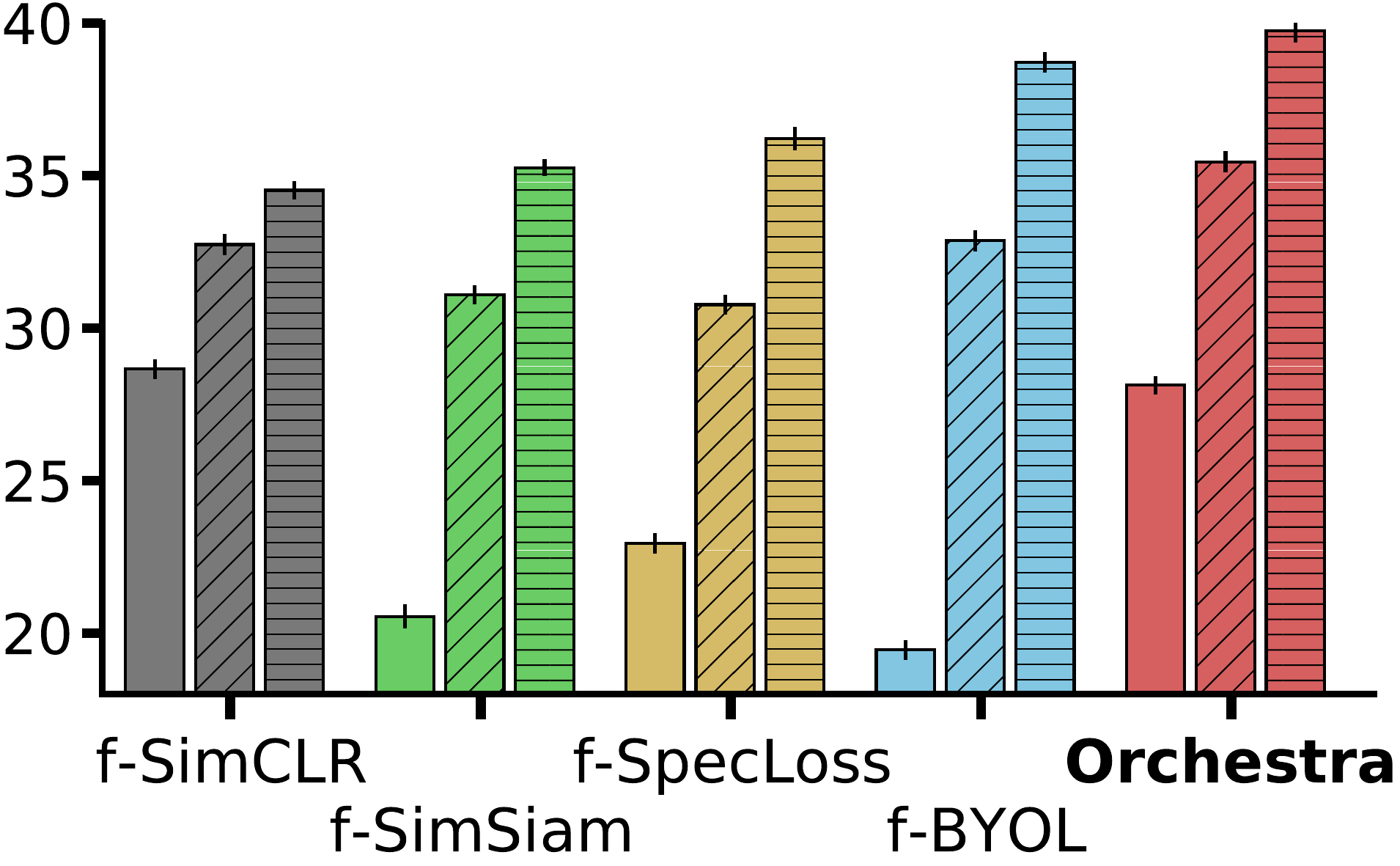}}
\end{subfigure}
\vspace{-2mm}
\caption{\label{fig:exptt_lepochs} \textbf{Robustness to local epochs} on CIFAR-10 (left) and CIFAR-100 (right). We find Orchestra achieves similar accuracy to other methods in half the number of local epochs.}
\vspace{-15pt}
\end{figure}

\textbf{Number of Clients.}
We next consider the effects of changing the number of clients. Following prior work~\cite{flcollab}, we linearly scale the number of local epochs to ensure that the total number of training iterations remains constant across all settings. Results are shown in \autoref{fig:exptt_scale}. We find that unlike other methods, which suffer from fluctuations in performance depending on number of clients, Orchestra achieves similar performance for all settings.

\textbf{Participation Ratio.}
Since only a fraction of participants can be expected to be connected to the server at a given time, participation ratio is an important attribute of cross-device FL. As shown in \autoref{fig:exptt_participation}, all methods except Orchestra suffer large decreases in accuracy when participation ratio is decreased. In contrast, Orchestra maintains accuracy even for the smallest participation ratios.

\textbf{Local Epochs.}
Resource constraints may force one to limit local training to few epochs, making robustness to limited epochs a valuable attribute. \autoref{fig:exptt_lepochs} shows that reducing local epochs causes all methods to lose accuracy, but Orchestra is the most resistant to this loss. For example, on CIFAR-10, Orchestra with 5 local epochs matches the performance of other methods with 10 local epochs, i.e., using Orchestra the training cost halves.

\begin{figure}
\centering
\begin{subfigure}{0.48\columnwidth}
  \centering
  \centerline{\includegraphics[width=\columnwidth]{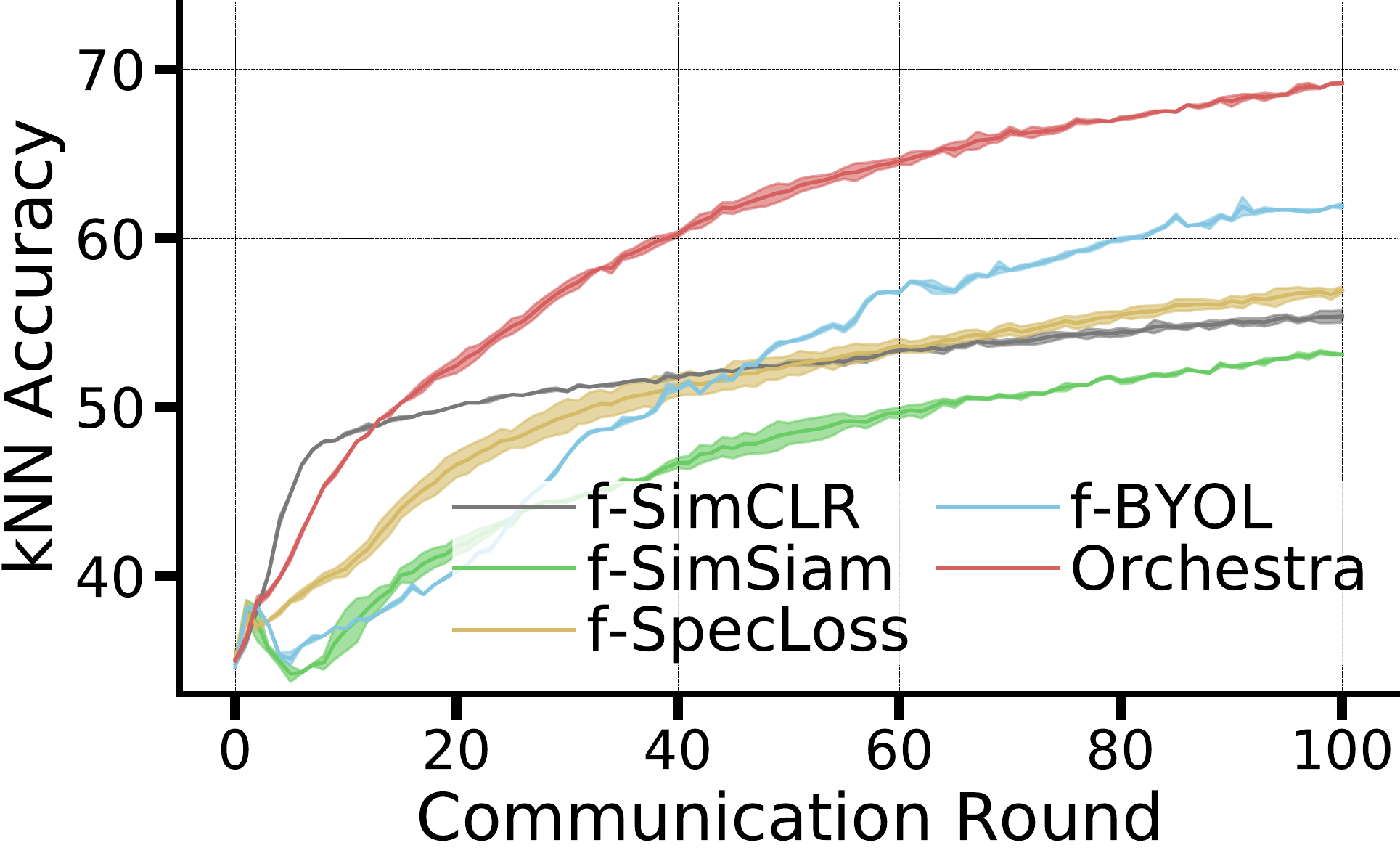}}
\end{subfigure}%
\begin{subfigure}{0.48\columnwidth}
  \centering
  \centerline{\includegraphics[width=\columnwidth]{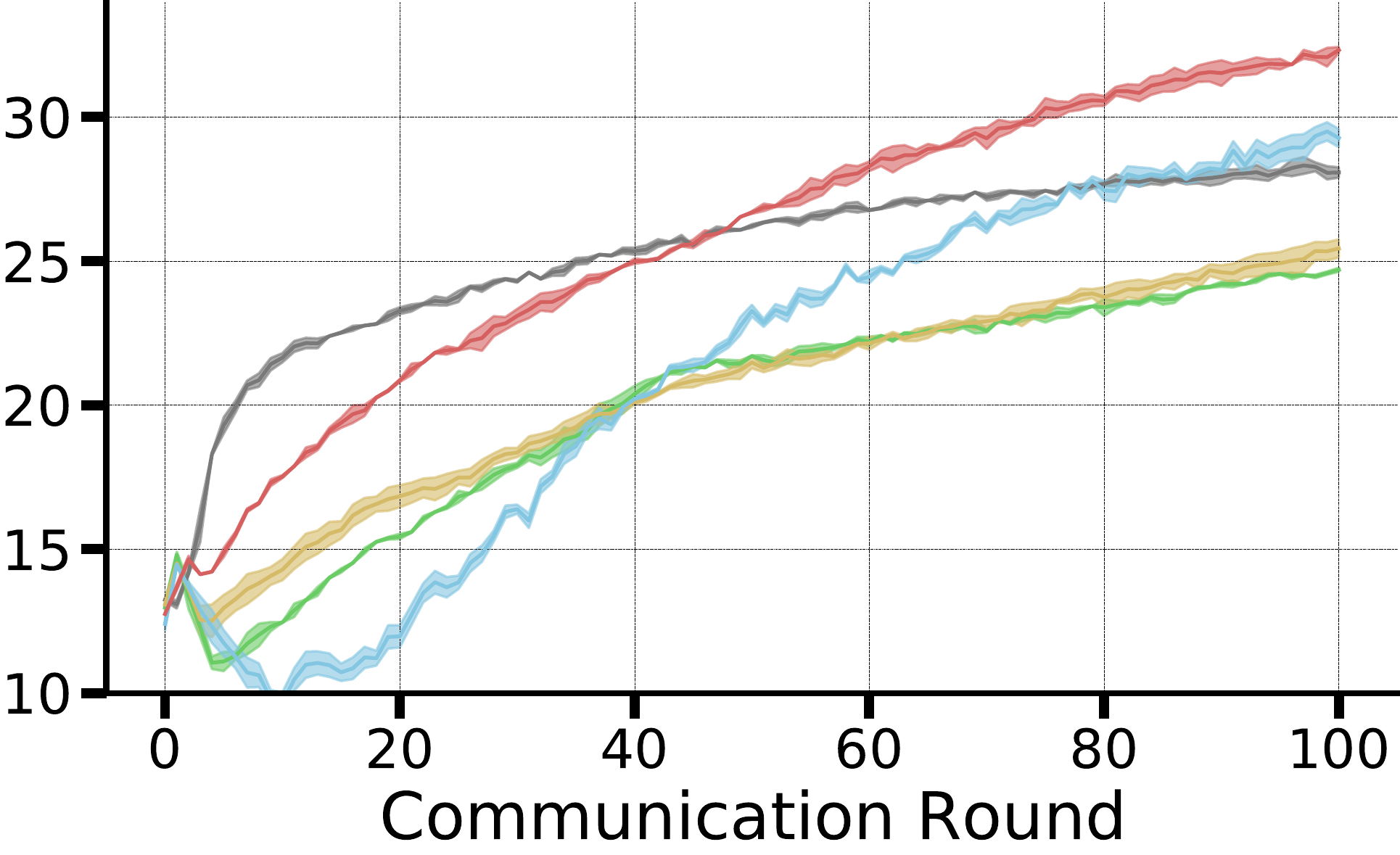}}
\end{subfigure}
\vspace{-2mm}
\caption{\label{fig:comms} \textbf{Communication efficiency} on CIFAR-10 (left) and CIFAR-100 (right). We plot kNN accuracy w.r.t.\ comm.\ rounds. Orchestra has the fastest round-to-accuracy response for moderate to high accuracies; SimCLR performs well on smaller accuracies.}
\vspace{-15pt}
\end{figure}

\textbf{Communication Efficiency.} Client-server communication is one of the major bottlenecks for FL in cross-device settings. Hence, it is ideal if an FL technique converges in fewer rounds. We measure the kNN accuracy of different methods during training and plot it as a function of the number of rounds. As shown in \autoref{fig:comms}, Orchestra takes the fewest rounds to achieve moderate to high accuracy; meanwhile, for smaller values, SimCLR is often the fastest to converge, but achieves poor ultimate accuracy.

\subsection{Attributes Specific to Orchestra} 

\begin{figure}
\centering
\begin{subfigure}{0.48\columnwidth}
  \centering
  \centerline{\includegraphics[width=\columnwidth]{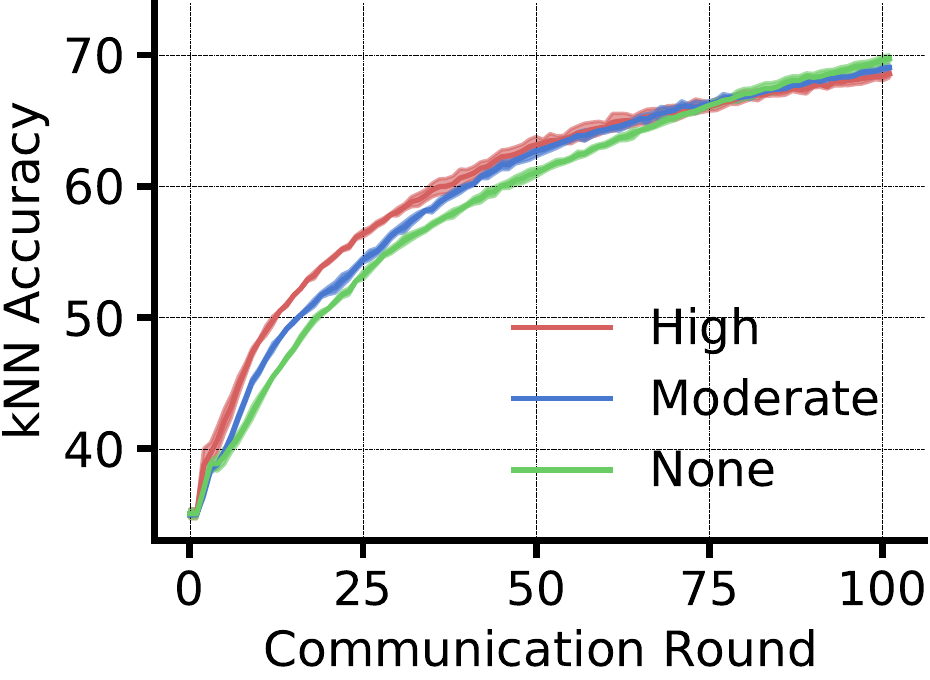}}
\end{subfigure}%
\begin{subfigure}{0.48\columnwidth}
  \centering
  \centerline{\includegraphics[width=\columnwidth]{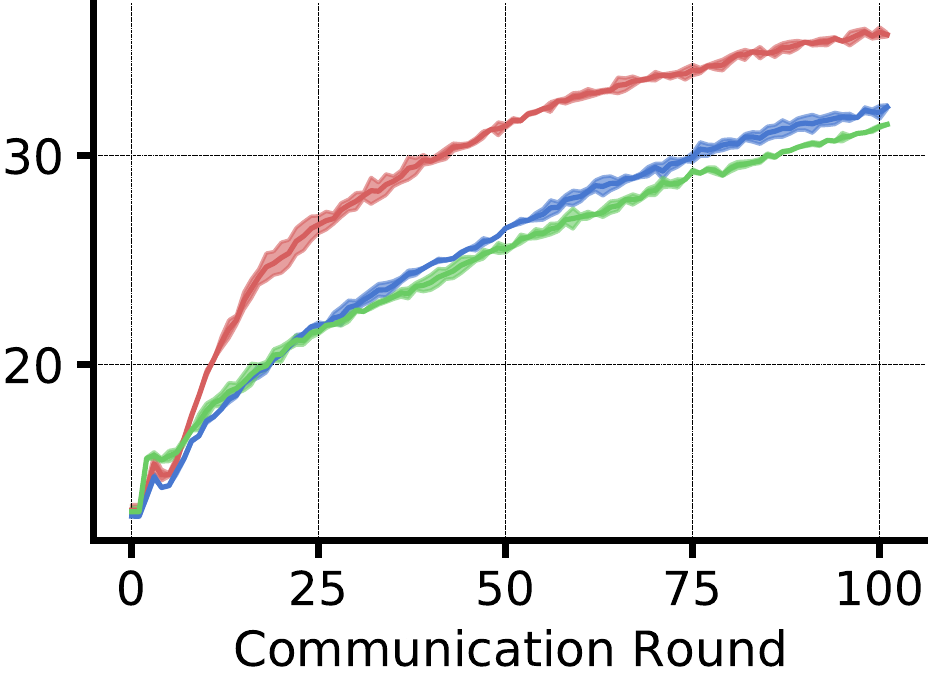}}
\end{subfigure}
\vspace{-2mm}
\caption{\label{fig:orchestra_comms} \textbf{Heterogeneity improves communication efficiency} for Orchestra on CIFAR-10 (left) and CIFAR-100 (right), i.e., Orchestra thoroughly exploits heterogeneity to maximize efficiency.}
\vspace{-0.5cm}
\end{figure}

\begin{table}
\caption{\label{tab:lg} \textbf{Sensitivity to number of global ($G$) and local ($L$) clusters.} We change $G$ and $L$ one-by-one, keeping the other fixed. Base values are reported in parentheses. ``Reps'' denotes the idealized setting, where representations are shared without local clustering. As shown, Orchestra has minimal sensitivity to both, the number of global and local clusters. Further, the performance is generally very close to the idealized setting.}
\scriptsize
\begin{tabular}{@{}c|cccc||cccc@{}}
\toprule
     & \multicolumn{4}{c||}{CIFAR-10 ($G$=32, $L$=8)}                                                                & \multicolumn{4}{c}{CIFAR-100 ($G$=256, $L$=16)}                                                                \\ \midrule
$G$    & \multicolumn{1}{c|}{8}     & \multicolumn{1}{c|}{16}    & \multicolumn{1}{c|}{64}    & 128    & \multicolumn{1}{c|}{64}    & \multicolumn{1}{c|}{128}   & \multicolumn{1}{c|}{256}   & 512   \\
Acc & \multicolumn{1}{c|}{70.25} & \multicolumn{1}{c|}{71.05} & \multicolumn{1}{c|}{71.04} & 71.38 & \multicolumn{1}{c|}{39.89} & \multicolumn{1}{c|}{40.27} & \multicolumn{1}{c|}{40.37} & 40.29 \\ \midrule 
$L$    & \multicolumn{1}{c|}{2}     & \multicolumn{1}{c|}{4}     & \multicolumn{1}{c|}{16}    & Reps     & \multicolumn{1}{c|}{8}     & \multicolumn{1}{c|}{16}    & \multicolumn{1}{c|}{64}    & Reps     \\
Acc & \multicolumn{1}{c|}{70.41} & \multicolumn{1}{c|}{71.06} & \multicolumn{1}{c|}{71.28} & 71.95 & \multicolumn{1}{c|}{40.09} & \multicolumn{1}{c|}{40.37} & \multicolumn{1}{c|}{40.25} & 41.19 \\
\bottomrule
\end{tabular}
\vspace{-0.5cm}
\end{table}

\textbf{Effect of Heterogeneity on Communication Efficiency.} Our theoretical result indicates and empirical results corroborate the expectation that Orchestra thrives under heterogeneity. We provide another demonstration of this result in \autoref{fig:orchestra_comms}, where we show the progress of kNN accuracy as a model is trained using Orchestra. We find that not only does heterogeneity help Orchestra improve performance, it can also improve its speed of optimization. 


\textbf{Number of Clusters.}
Orchestra's underlying principles are rooted in Prop.~\ref{prop:distcluster} and \ref{prop:fedcluster}, which provide the goal of finding discriminable clusters. However, another important attribute in those bounds is the number of global clusters $G$. If the $G$ is smaller than the number of classes, generalization can suffer. Similarly, having few local clusters $L$ can lead to inconsistent assignments (larger $c$) and hurt generalization. More local clusters can help avoid this, but if the number approaches the size of the local dataset, privacy (anonymity) is lost because all clusters will contain only one datapoint. We thus study sensitivity of Orchestra to $G$ and $L$. \autoref{tab:lg} shows that Orchestra is robust to the number of global clusters as long as it even slightly exceeds the number of classes, achieving similar performance in all settings. We similarly see that Orchestra is essentially robust to the number of local centroids, but can see minimal performance loss if very few clusters are used. Finally, these results also indicate the values of $G$ and $L$ need not be precisely tuned, as most settings yield good performance.

\section{Conclusion}
\label{sec:conclusion}
\vspace{-0.1cm}
To enable wider adoption of federated learning (FL) for complex modalities such as vision, we need to reduce its reliance on labeled data. Towards this goal, we presented \emph{Orchestra}, an unsupervised FL technique that orchestrates a distributed clustering task and enforces a globally consistent partitioning of clients' data, while remaining mindful of the core challenges seen in FL setups. Built on strong theoretical foundations, Orchestra is scalable, achieves strong communication-efficiency, thrives under heterogeneity, and remains robust to various FL parameters.

\section*{Acknowledgements}
We thank the teams at Nokia Bell Labs, UK and Flower.dev for several useful discussions during the course of this project. We also thank the reviewers and AC for multiple useful suggestions that helped improve the paper. ESL's time at University of Michigan was partly supported via NSF award CNS-2008151.

\bibliography{main}
\bibliographystyle{icml2022}

\newpage
\appendix
\onecolumn
\begin{center} 
{\Large \textbf{Appendix}}
\end{center} 

In this appendix, we provide more description about the federated learning settings studied in this paper (\S\ref{sec:cross-fl}), and contextualize prior works based on FL properties studied in them (\S\ref{sec:more_rw}). Thereafter, we provide the formal algorithm for Orchestra and elaborate on our experiment, training, and evaluation setups (\S\ref{appendix:experiment}). Next, \S\ref{appendix:results} contains T-SNE visualizations, ablation results, and shows the performance of Orchestra under a state-of-the-art differentially private local clustering technique~\cite{dpchang}. The appendix ends by providing a recap of the key notations used in the paper and proofs for Propositions~\ref{prop:distcluster} and \ref{prop:fedcluster}. 

\section{Properties of Cross-Silo and Cross-Device FL}
\label{sec:cross-fl}
Federated learning setups vary across different applications and circumstances. In general, they can be categorized into two main settings: \emph{Cross-silo} and \emph{Cross-device}, as described in a recent federated learning survey by~\cite{flsurvey}. 

The cross-silo setting shares several properties of datacenter distributed learning, where the number of clients is often limited, and the clients have a large amount of data and high-level of computational resources, are stateful, almost always available with few failures. However, these requirements often are not applicable to many real-life FL setups, and the cross-silo setting is more relevant when large organizations collaborate. 

On the other hand, the cross-device setting has much fewer requirements on the clients: they can be any number of mobile or IoT devices which have limited computational resources, be stateless, unavailable at any time, and unreliable. The relaxation on the requirement of clients make it more applicable to real-world FL setups, but any proposed solution must tackle the challenges that this setting brings.

Many existing centralized unsupervised representation learning algorithms are relatively straight-forward to be adapted to work in a cross-silo setup, where the clients are often computationally powerful and can be assumed as stateful. However, clients in the cross-device setting often have much less resources, and this brings many challenges in adapting centralized algorithms. For example, the requirement of large batch-sizes~\cite{ssfl} is difficult to meet in the cross-device setting as mobile or IoT devices have limited runtime memory. Similarly, sharing representations across clients as done in ~\cite{fedmoco, fedca} is disallowed due to user privacy concerns. Hence, designing a method which scales to hundreds of participating clients, is stateless, works with small batch sizes and uneven distribution of data, and preserves user privacy is crucial to the success of \emph{cross-device} unsupervised representation learning. This was the primary motivation behind the design of Orchestra.

\begin{table}[h]
\centering
\begin{tabular}{@{}ccccccc@{}}
\toprule
                                                                     & \multicolumn{6}{c}{\textbf{Properties}}                                                                                                                                      \\ \midrule
\textbf{Methods}                                                     & Stateless  & \begin{tabular}[c]{@{}c@{}}Privacy\end{tabular} & Supports small  & \# Clients     & Cross-silo & Cross-Device \\ 
                                                     &    & \begin{tabular}[c]{@{}c@{}} Preserving\end{tabular} & Batch Size  &      &  &  \\ \hline
FedCA~\cite{fedca}                                                                & \checkmark          &              $\times$                                                                                       &     $\times$ {\scriptsize (128)}    & 5               & \checkmark          &      $\times$        \\
FedU~\cite{flcollab}                                                                 &  $\times$          & \checkmark                                                                                                   &      $\times$ {\scriptsize (128)}    & 5               & \checkmark          &       $\times$       \\
FedEMA~\cite{fldivssl}                                                               & $\times$           & \checkmark                                                                                                   &  $\times$ {\scriptsize (128)}        & 5               & \checkmark          &       $\times$       \\
SSFL~\cite{ssfl}                                                                 & \checkmark          & \checkmark                                                                                                   &     $\times$ {\scriptsize (256)}    & 10              & \checkmark          &      $\times$        \\
FCL~\cite{wu2022federated}             & \checkmark          & \begin{tabular}[c]{@{}c@{}}Using an additional \\ encryption module\end{tabular}        &    $\times$ {\scriptsize (128)}   & 5-10         & \checkmark     &    $\times$          \\\hline
\textbf{Orchestra}                                                   & \textbf{\checkmark } & \textbf{\checkmark }                                                                                          & \textbf{\checkmark} {\scriptsize (16)}  & \textbf{10-400} & \textbf{\checkmark } & \textbf{\checkmark }   \\ \bottomrule
\end{tabular}
\caption{Comparison of Orchestra with prior federated unsupervised learning approaches. Orchestra is unique in its capability to learn from unlabeled data in a small batch size regime without requiring any stateful operations or sharing local representations with the server. Moreover, unlike prior approaches which were evaluated only in small scale setups with 5-10 clients, Orchestra can easily scale to large-scale cross-device settings with hundreds of participating clients.}
\label{table:rw}
\end{table}

\section{More Related Work}
\label{sec:more_rw}
In Table~\ref{table:rw}, we list various unsupervised FL approaches proposed in the recent literature and compare their properties with Orchestra. As evident, Orchestra is unique in its capability to learn from unlabeled data in a \emph{small batch size} regime without requiring any stateful operations or sharing local representations with the server. Moreover, unlike prior approaches which were evaluated only in small scale setups with 5-10 clients, Orchestra can easily scale to large-scale settings with hundreds of participating clients, making it particularly apt for \emph{cross-device} FL.

\textbf{Self-Supervised Learning:} SSL techniques are a recent paradigm in centralized unsupervised learning, wherein a task-relevant signal is extracted from the data itself and used to guide the training of a model. In vision, specifically, this task-relevance prior is encoded by designing pretext tasks. Earlier examples of such tasks include predictive tasks, such as rotation prediction~\cite{rotnet} or predicting the patch order in a shuffled image~\cite{jigsaw}. Recently, though, the task of instance discrimination has revolutionized unsupervised visual training~\cite{instancedisc}. Herein, task-relevant information in encoded by enforcing invariances to a set of data augmentations~\cite{demystify, featurelearning, stylecontent}. Popular methods include contrastive techniques (SimCLR~\cite{simclr}, MoCo~\cite{moco}, SpecLoss~\cite{specloss}); similarity-based techniques (BYOL~\cite{byol}, SimSiam~\cite{simsiam}, DINO~\cite{dino}); redundancy reduction methods (Barlow Twins~\cite{btwins}, VICReg~\cite{vicreg}); and clustering based methods (SeLa~\cite{sela}, SwAV~\cite{swav}, PCL~\cite{pcl}).

\section{Experimental Details}
\label{appendix:experiment}

\subsection{Data}

\textbf{Datasets.} Our experiments focus on the widely-used CIFAR-10 and CIFAR-100 datasets~\cite{krizhevsky2009learning}. Both datasets consist of 60,000 images, divided into two partitions: 50,000 for training and 10,000 for testing. CIFAR-10 consists of 10 object classes whereas CIFAR-100 have 100 object classes; with training samples equally distributed across all classes. For federated training, we partitioned both the datasets across $K$ clients. To this end, we sample class priors from a Dirichlet distribution~\cite{dirchsplit} controlled by the Dirichlet parameter $\alpha$. A smaller value of $\alpha$ yields more non-IID splits across clients. More specifically, we experimented with three different levels of heterogeneity by setting $\alpha$ to 10$^{\text{5}}$ (IID), 10$^{\text{-1}}$ (moderately non-IID), and 10$^{\text{-3}}$ (highly non-IID). To quantify the level of heterogeneity, we provide average number of classes in \autoref{tab:nclass}. We consider two definitions for a class to be present on a client: if at least 1 sample from the class is present on the client and if at least 1\% samples belong to the class. $\sim$ denotes an experiment that was not conducted.

\textbf{Transformations.} During the local training stage shown in Figure~\ref{fig:orchestra_all}, we need to generate an augmentation of each input sample. We use the same set of augmentations proposed by Grill et al.~\yrcite{byol}. For the baseline SSL methods, we follow the augmentations proposed in their original papers.

\begin{table}[h]
\centering
\caption{\label{tab:nclass} \textbf{Average number of classes per client.} We use three values of $\alpha$: 10$^{\text{5}}$ (IID), 10$^{\text{-1}}$ (moderately non-IID), and 10$^{\text{-3}}$ (highly non-IID). We define a class to be present on a client in two ways: (i) at least 1 sample from the class is present on the client; (ii) at least 1\% samples on the client belong to the class. $\sim$ denotes an experiment that was not conducted.}
\begin{tabular}{@{}c|c|ccc|ccc@{}}
\toprule
                                                                                      &          & \multicolumn{3}{c|}{CIFAR-10} & \multicolumn{3}{c}{CIFAR-100} \\ \midrule
                                                                                      & $\alpha$         & $10^{-3}$     & $10^{-1}$    & $10^{5}$   & $10^{-3}$    & $10^{-1}$     & $10^{5}$   \\
                                                                                      & Heterogeneity         & High     & Moderate    & None   & High    & Moderate     & None   \\ \midrule
\multirow{2}{*}{\begin{tabular}[c]{@{}c@{}}Cross-Device\\ (100 clients)\end{tabular}} & 1 sample & 1.08      & 4.69   & 10       & 3.56     & 29.3    & 100      \\
                                                                                      & 1 \%     & 1.05      & 3.13   & 10       & 2.13     & 17.92   & 100      \\ \midrule
\multirow{2}{*}{\begin{tabular}[c]{@{}c@{}}Cross-Silo\\ (10 clients)\end{tabular}}    & 1 sample & $\sim$    & 6.25   & 10       & $\sim$   & 48.63   & 100      \\
                                                                                      & 1 \%     & $\sim$    & 4.87   & 10       & $\sim$   & 35.5    & 100      \\ \bottomrule
\end{tabular}
\end{table}

\subsection{Algorithm, Implementation, and Training Details}
\label{subsec:algo}

\textbf{Algorithm:} Below we provide a detailed algorithm of our pipeline, as described in \S\ref{sec:method} and outlined in \autoref{fig:orchestra_all}. Orchestra involves federation and communication between clients and the server, and \autoref{alg:orchestra_local} describes local training happening on the clients, while \autoref{alg:orchestra_server} outlines the computation happening on the server. We refer to the backbone and projector models jointly as the `feature encoder'. Broadly, each client trains its feature encoders locally with both clustering and degeneracy losses, and returns the local clusters. The server aggregates the feature encoders from all clients using any federated averaging algorithm (e.g., FedAvg), and performs further clustering on the local centroids returned by all the clients to obtain global centroids. The global centroids are then passed back to the clients for another round of local training.

When Orchestra starts, the server has to initialize the global centroids to be used for training (Line 2 of \autoref{alg:orchestra_server}). However, as the server does not have any prior data about the clients, the initial global centroids are initialized with a small federation step with the clients: the initialized feature encoder $f_T^G$ is passed to the clients, and the clients compute the representations by passing the local data through the encoder (similar to Line 5 of \autoref{alg:orchestra_local}). These representations are then clustered (Line 18 of \autoref{alg:orchestra_local}) and the centroids are sent to the server, where a further clustering is performed (Line 8 of \autoref{alg:orchestra_server}), forming the initial global centroids $\mu^G$.

The clustering algorithm $\mathcal{C}$ used in both the local and global clustering processes is based on the Sinkhorn-knopp algorithm~\cite{otclustering}. We run local clustering on a memory module that stores 128 most recent representations computed during local training, i.e., representations computed during last 8 iterations. This allow us to avoid the cost of computing model representations for local clustering again. Since we use the target model for clustering purposes, these representations can be expected to be essentially the same as the representations that the model with latest parameters would compute. Combined, this trick makes local clustering \emph{very efficient, with no extra inference/memory costs}. 
\setlength{\intextsep}{2pt}%
\setlength{\columnsep}{7pt}%
\begin{wrapfigure}{r}{0.35\textwidth}
  \vspace{-5pt}
  \begin{center}
    \includegraphics[width=0.35\textwidth]{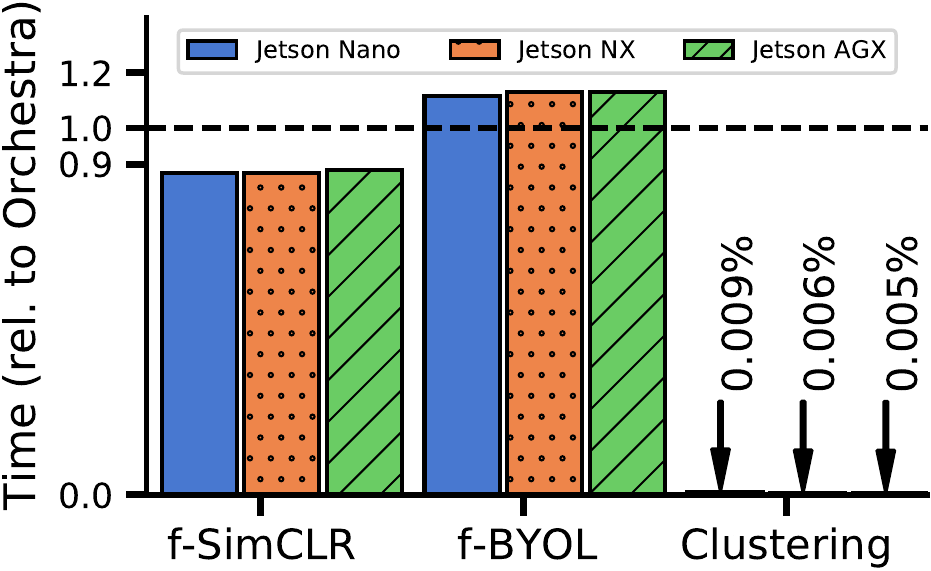}
  \end{center}
  \vspace{-10pt}
  \caption{\label{fig:time}Client runtime per round, relative to Orchestra.}
\end{wrapfigure}
In fact, we demonstrate this in \autoref{fig:time} by successfully deploying Orchestra on three NVIDIA Jetson embedded devices (with RAM as low as 4GB). We plot time consumed by a client running other methods, relative to when it runs Orchestra, and percent time consumed by local clustering in a round of Orchestra. As can be seen, Orchestra's latency per round is similar to other methods, and clustering accounts for $\leq0.009$\% of the training cost, \emph{confirming Orchestra's practicality for cross-device FL}.

\textbf{Implementation and Code:} Orchestra is implemented using PyTorch and the Flower federated learning framework~\cite{beutel2020flower}. Our primary results in cross-device FL settings are presented for $K=100$ clients, which we simulate on 8 NVIDIA V100 GPUs using Flower's Virtual Client Engine. Consistent with the paradigm of cross-device FL, we use a small batch size of 16 on each client, set the number of local epochs $E$ to 10, communication rounds to {100}, and participation ratio to 0.5. For cross-silo experiments, a batch size of 256 and participation ratio of 1.0 is employed. While evaluating the scalability and communication efficiency of Orchestra in cross-device settings (\S\ref{sec:fl_experiments}), we also show results for different values of $K$ and $E$. 

A working source code of Orchestra is provided available at following \href{https://github.com/akhilmathurs/orchestra}{\color{red}{github link.}}

\begin{algorithm}[!thb]
    \caption{Orchestra - Local Training}
    \label{alg:orchestra_local}
\begin{algorithmic}[1]
    \STATE {\bfseries Require:} Data $X^k$ on client $k$, global centroids $\mu^G$, batch size $N$, exponential update rate $m$, target encoder $f_{\text{T}}$, online encoder $f_{\text{O}}$, stochastic augmentation function $\mathcal{T}$, rotation function $\mathcal{R}$, clustering function with size constraints $\mathcal{C}$
    \FOR{sampled minibatch $\{x\}^N_{i=1} \subset X^k$}
        \FOR{$i \in \{1,\ldots,N\}$} 
            \STATE Compute transformed sample: $\tilde{x_i} \gets \mathcal{T}(x_i)$
            \STATE Compute representations: $f_{\text{T}}(x_i)$ $\,$and$\,$ $f_{\text{O}}(\tilde{x_i})$
            \STATE Compute cluster assignment according to global centroids: $P_{f_{\text{T}}}(x_i)$ and $P_{f_{\text{O}}}(\tilde{x_i})$
            \STATE Select an random rotation angle index: $a_i \sim  \mathcal{U}_{4}$
            \STATE Convert to one hot encoding: $R_i \gets \text{onehot}(a_i) $
            \STATE Rotate sample: $\hat{x_i} \gets \mathcal{R}(x_i, \alpha_{a_i})$
            \STATE Compute representation: $f_{\text{O}}(\hat{x_i})$
            \STATE Predict rotation: $\hat{R_i} \gets \text{softmax} \left(W_{r}^{T}f_O(\hat{x_i})\right)$
        \ENDFOR
        \STATE Compute clustering loss: $\mathcal{L}_{\text{cluster}} = - \frac{1}{N} \sum^N_{i=1} (\mathcal{H}\left(P_{f_\text{T}}(x_i), P_{{f_\text{O}}}(\tilde{x_i})\right))$
        \STATE Compute degeneracy loss: $\mathcal{L}_{\text{deg}} = - \frac{1}{N} \sum^N_{i=1} (\mathcal{H}(R_i, \hat{R_i}))$
        \STATE Update $f_{\text{O}}$ to minimize $\mathcal{L} = \mathcal{L}_{\text{cluster}} + \mathcal{L}_{\text{deg}}$
        \STATE Update $f_{\text{T}} \gets m f_{\text{T}} + (1 - m)
        f_{\text{O}}$
        
    \ENDFOR
    \STATE Cluster $\mu^{L}_k \gets \mathcal{C}(\{f_{\text{T}}(x) | x \in X^k\}, L^{(k)})$
    \STATE {\bfseries return} trained encoders $f_{\text{T}}$, $f_{\text{O}}$, local clusters $\mu^{L}_k$
\end{algorithmic}
\end{algorithm}

\begin{algorithm}[!thb]
    \caption{Orchestra - Server Federation}
    \label{alg:orchestra_server}
\begin{algorithmic}[1]
    \STATE {\bfseries Require:} Number of rounds $N$, Number of global centroids $G$, clustering function with size constraints $\mathcal{C}$
    \STATE Initialize global target encoder $f_{\text{T}}^G$, global online encoder $f_{\text{O}}^G$, global centroids $\mu^G$
    \FOR{round $i \in \{1,\ldots,N\}$}
        \STATE Orchestrate local training with $f_{\text{T}}^G, f_{\text{O}}^G, \mu^G$
        \STATE Collect results: $\{(f_{\text{T}}^k$, $f_{\text{O}}^k, \mu^{L}_k) | k \in [K] \}$
        \STATE Aggregate encoders: $f_{\text{T}}^G \gets \frac{1}{K} \sum^K_{i=1} f_{\text{T}}^k$ $\,$ and $\,$ $f_{\text{O}}^G \gets \frac{1}{K} \sum^K_{i=1} f_{\text{O}}^k$
        \STATE Aggregate centroids: $\mu^L \gets \{\mu | k \in [K], \mu \in \mu^{L}_k\}$
        \STATE Update global centroids by clustering: $\mu^{G} \gets \mathcal{C}(\mu^L, G)$
    \ENDFOR
    \STATE {\bfseries return} trained encoder $f_{\text{T}}^G$
\end{algorithmic}
\end{algorithm}

\subsection{Network Architectures}
Orchestra uses ResNet-18 as the network architecture for the Backbone, and a 2-layer multi-layer perceptron (MLP) with 512 units in each layer for the Projector, following SimCLR~\cite{simclr}. No Predictor network is used. For all the baselines, we use ResNet-18 as the Backbone, while Projector and Predictor architectures are borrowed from their respective papers. 

\subsection{Hyperparameter Tuning and Scaling Schemes}
\textbf{Tuning:} As mentioned in \autoref{sec:tuning}, we find configuration that maximize a linear combination of alignment and uniformity scores (see \autoref{eq:algn_unif}). We set $\tau$ to 0.2, similar to \autoref{fig:tuning}. Tuning is performed by running a model for 20 communication rounds with 10 local epochs. Other settings such as batch-size and participation ratio depend on whether we are in a cross-silo or a cross-device setup and are defined above. In the following, we report our hyperparameter grids and the retrieved values from tuning different methods. We denote learning rate using $\eta$ and EMA value using $m$.
\begin{enumerate}
    \item SimCLR:  $\eta$: \{0.03 (Cross-silo), 0.01, 0.003 (Cross-device), 0.001\}
    \item SimSiam:  $\eta$: \{0.03 (Cross-silo), 0.01 (Cross-device), 0.003, 0.001\}
    \item SpecLoss: $\eta$: \{0.03 (Cross-silo), 0.01, 0.003 (Cross-device), 0.001\}
    \item BYOL: $\eta$: \{0.03 (Cross-silo), 0.01 (Cross-device), 0.003, 0.001\}; $m$: \{0.9, 0.99 (Cross-silo), 0.996 (Cross-device)\}
    \item Orchestra:  $\eta$: \{0.03, 0.01 (Cross-silo), 0.003 (Cross-device), 0.001\}; $m$: \{0.9, 0.99 (Cross-silo), 0.996 (Cross-device)\}
\end{enumerate}

\textbf{Scaling Schemes for number of clients and participation ratio experiments:} We tune our methods for two baseline settings: (i) Cross-device with 100 clients; (ii) Cross-silo with 10 clients. For experiments where we change other number of clients and participation ratio, we follow previous works and scale number of local epochs or learning rate to avoid the costs of tuning a method again. Specifically, we have the following schemes.
\begin{itemize}
    \item \textbf{Number of Clients:} When varying number of clients, we follow Zhuang et al.~\yrcite{flcollab} and linearly scale number of local epochs. Given other variables remain fixed, this ensures a constant training budget in terms of number of iterations. For example, if $L$ is our base learning rate for a setting with $K$ clients, we scale the number of local epochs for a setting with $K_{\text{new}}$ clients as follows: $L_{\text{new}} = L \cdot \frac{K_{\text{new}}}{K}$.
    \item \textbf{Participation Ratio:} When varying participation ratio, we follow Charles et al.~\yrcite{largecohort} and quadratically scale the learning rate. Given other variables remain fixed, this helps achieve consistent performance across a large number of settings for number of clients. For example, if $\eta$ is our base learning rate for a setting with $R$ participation ratio, we scale the learning rate for a setting with $R_{\text{new}}$ participation ratio as follows: $\eta_{\text{new}} = \eta \cdot \sqrt{\frac{R_{\text{new}}}{R}}$.
\end{itemize}

\subsection{Evaluation Protocol}
To evaluate the quality of the representations learned by Orchestra, we primarily use the standard linear probe protocol, where the model is frozen and a linear classifier is learned on top of the backbone~\cite{simclr}. When comparing rounds to accuracy, we also use kNN accuracy probe~\cite{simsiam}. Finally, during semi-supervised evaluation, we fine-tune the entire model under limited labelled data (1\% or 10\% labels) where is held-out during the unsupervised training stage. 


\section{Additional Results}
\label{appendix:results}

\subsection{Linear and Semi-Supervised Evaluation Results (with standard deviations)}

In Table~\ref{tab:appendix_full_table}, we report the mean and standard deviation of accuracies obtained using linear and semi-supervised Evaluation. Please note that this Table~\ref{tab:appendix_full_table} is an extension of Table~\ref{tab:eval_results} presented in the main paper, with standard deviation values added to it. Each experiment was run three times with different seeds to obtain the mean and standard deviation scores.  

\begin{table*}[t]
\caption{\label{tab:appendix_full_table} We analyze Orchestra in the cross-device and cross-silo setting on CIFAR-10/CIFAR-100 datasets. For cross-device settings, due to a lack of baselines, we use FL-extensions of several centralized techniques; for cross-silo settings, we follow implementations of these techniques proposed by recent works that use stateful clients and divergence-aware predictor updates~\cite{flcollab}. We evaluate models using the popular linear probe technique~\cite{simclr} and semi-supervised fine-tuning with 1\% and 10\% labelled data.}
\begin{subtable}[h]{\textwidth}
\small
\centering
\begin{tabular}{@{}c|cccccc@{}}
\toprule
Dataset      & \multicolumn{6}{c}{CIFAR-10}                                                                                                                                                                          \\ \midrule
Setting      & \multicolumn{3}{c|}{Cross-Device (K = 100)}                                                                               & \multicolumn{3}{c}{Cross-Silo (K=10)}                                                                                          \\ \midrule
	         & \multicolumn{1}{c|}{Linear}         & \multicolumn{1}{c|}{1\%}            & \multicolumn{1}{c|}{10\%}           & \multicolumn{1}{c|}{Linear}         & \multicolumn{1}{c|}{1\%}            & 10\%           \\ \midrule
f-SimCLR       & \multicolumn{1}{c|}{58.36 $\pm$ 0.19}          & \multicolumn{1}{c|}{41.95 $\pm$ 0.85}          & \multicolumn{1}{c|}{44.64 $\pm$ 0.71}          & \multicolumn{1}{c|}{69.29 $\pm$ 0.28}          & \multicolumn{1}{c|}{57.76 $\pm$ 0.33}          & 68.27  $\pm$ 0.67     \\
f-SimSiam      & \multicolumn{1}{c|}{61.61 $\pm$ 0.68}          & \multicolumn{1}{c|}{49.99 $\pm$ 0.26}          & \multicolumn{1}{c|}{56.43 $\pm$ 0.52}          & \multicolumn{1}{c|}{75.12 $\pm$ 0.38}          & \multicolumn{1}{c|}{64.04 $\pm$ 0.41}          & 72.25 $\pm$ 0.37          \\ 
f-SpecLoss     & \multicolumn{1}{c|}{66.51 $\pm$ 0.53}          & \multicolumn{1}{c|}{55.66 $\pm$ 0.80}          & \multicolumn{1}{c|}{62.09 $\pm$ 0.67}          & \multicolumn{1}{c|}{{\ul 80.71 $\pm$ 0.31}}    & \multicolumn{1}{c|}{70.88 $\pm$ 0.53}          & {\ul 77.96 $\pm$ 0.55}    \\
f-BYOL         & \multicolumn{1}{c|}{{\ul 65.85 $\pm$ 0.19}}    & \multicolumn{1}{c|}{{\ul 56.05 $\pm$ 0.31}}    & \multicolumn{1}{c|}{{\ul 64.15 $\pm$ 0.30}}    & \multicolumn{1}{c|}{76.08 $\pm$ 0.30}          & \multicolumn{1}{c|}{{\ul 65.55 $\pm$ 0.34}}    & 73.18 $\pm$ 0.40           \\
Orchestra    & \multicolumn{1}{c|}{\textbf{71.58 $\pm$ 0.53}} & \multicolumn{1}{c|}{\textbf{60.33 $\pm$ 0.63}} & \multicolumn{1}{c|}{\textbf{66.20 $\pm$ 0.71}} & \multicolumn{1}{c|}{\textbf{82.14 $\pm$ 0.38}} & \multicolumn{1}{c|}{\textbf{71.30 $\pm$ 0.27}} & \textbf{79.51 $\pm$ 0.51} \\\midrule
RotPred (pred) & \multicolumn{1}{c|}{44.44 $\pm$ 0.93}          & \multicolumn{1}{c|}{34.71 $\pm$ 0.69}          & \multicolumn{1}{c|}{46.15 $\pm$ 0.65}          & \multicolumn{1}{c|}{55.68 $\pm$ 0.38}          & \multicolumn{1}{c|}{45.84 $\pm$ 0.53}          & 51.32 $\pm$ 0.55            \\
FedAvg (sup) & \multicolumn{1}{c|}{80.85 $\pm$ 0.37}          & \multicolumn{1}{c|}{82.76 $\pm$ 0.71}          & \multicolumn{1}{c|}{80.34 $\pm$ 0.61}          & \multicolumn{1}{c|}{79.22 $\pm$ 0.38}          & \multicolumn{1}{c|}{86.81 $\pm$ 0.53}          & 87.12 $\pm$ 0.77       \\\bottomrule
\end{tabular}
\end{subtable}

\vspace{3mm}

\begin{subtable}[h]{\textwidth}
\small
\centering
\begin{tabular}{@{}c|cccccc@{}}
\toprule
Dataset      & \multicolumn{6}{c}{CIFAR-100}                                                                                                                                                                          \\ \midrule
Setting      & \multicolumn{3}{c|}{Cross-Device (K = 100)}                                                                               & \multicolumn{3}{c}{Cross-Silo (K=10)}                                                                                          \\ \midrule
	         & \multicolumn{1}{c|}{Linear}         & \multicolumn{1}{c|}{1\%}            & \multicolumn{1}{c|}{10\%}           & \multicolumn{1}{c|}{Linear}         & \multicolumn{1}{c|}{1\%}            & 10\%           \\ \midrule
f-SimCLR       &  \multicolumn{1}{c|}{34.52 $\pm$ 0.34}          & \multicolumn{1}{c|}{45.47  $\pm$ 0.32}          & \multicolumn{1}{c|}{51.88  $\pm$ 0.56}          & \multicolumn{1}{c|}{44.33  $\pm$ 0.33}          & \multicolumn{1}{c|}{57.61  $\pm$ 0.27}          & 67.84   $\pm$ 0.15        \\
f-SimSiam & \multicolumn{1}{c|}{34.96  $\pm$ 0.43}          & \multicolumn{1}{c|}{47.17  $\pm$ 0.73}          & \multicolumn{1}{c|}{55.13  $\pm$ 0.41}          & \multicolumn{1}{c|}{43.16  $\pm$ 0.32}          & \multicolumn{1}{c|}{53.38  $\pm$ 0.66}          & 63.19   $\pm$ 0.81        \\ 
f-SpecLoss     & \multicolumn{1}{c|}{37.60  $\pm$ 0.37}          & \multicolumn{1}{c|}{47.11  $\pm$ 0.56}          & \multicolumn{1}{c|}{50.91  $\pm$ 0.15}          & \multicolumn{1}{c|}{\textbf{56.59  $\pm$ 0.45}} & \multicolumn{1}{c|}{{\ul 62.15  $\pm$ 0.67}}    & {\ul 72.09  $\pm$ 0.56}    \\
f-BYOL         & \multicolumn{1}{c|}{38.47  $\pm$ 0.34}          & \multicolumn{1}{c|}{{\ul 52.89  $\pm$ 0.62}}    & \multicolumn{1}{c|}{{\ul 58.56  $\pm$ 0.92}}    & \multicolumn{1}{c|}{49.64  $\pm$ 0.44}          & \multicolumn{1}{c|}{57.34  $\pm$ 0.45}          & 66.13   $\pm$ 0.63        \\
Orchestra   & \multicolumn{1}{c|}{\textbf{40.37  $\pm$ 0.30}} & \multicolumn{1}{c|}{\textbf{54.01  $\pm$ 0.41}} & \multicolumn{1}{c|}{\textbf{59.07  $\pm$ 0.69}} & \multicolumn{1}{c|}{{\ul 55.89  $\pm$ 0.49}}          & \multicolumn{1}{c|}{\textbf{63.73  $\pm$ 0.28}} & \textbf{73.06  $\pm$ 0.67} \\\midrule
RotPred (pred) &  \multicolumn{1}{c|}{16.85  $\pm$ 0.74}          & \multicolumn{1}{c|}{15.79  $\pm$ 0.64}          & \multicolumn{1}{c|}{19.52  $\pm$ 0.73}          & \multicolumn{1}{c|}{25.0  $\pm$ 0.39}           & \multicolumn{1}{c|}{27.64  $\pm$ 0.95}          & 28.81  $\pm$ 0.82         \\
FedAvg (sup) & \multicolumn{1}{c|}{58.71  $\pm$ 0.42}          & \multicolumn{1}{c|}{59.07  $\pm$ 0.50}          & \multicolumn{1}{c|}{64.01  $\pm$ 0.53}          & \multicolumn{1}{c|}{65.59  $\pm$ 0.38}          & \multicolumn{1}{c|}{62.48  $\pm$ 0.42}          & 71.59  $\pm$ 0.79         \\\bottomrule

\end{tabular}
\end{subtable}

\end{table*}

\begin{table}[b]
\centering
\small
\begin{tabular}{@{}cccc@{}}
\toprule
         & Base Setting & No Degeneracy Regularization & No Target Model \\ \midrule
CIFAR-10 &  71.58  & 56.62 & 68.63\\
CIFAR-100 & 40.37  &  28.86 & {36.8}\\
\bottomrule
\end{tabular}
\caption{\label{tab:ablations} Ablation results for Orchestra's target model and degeneracy regularization. These results show both solutions, degeneracy regularization and use of target model are important. However, noisy assignments can be overcome with time and hence sensitivity to use of target model is lower than the use of degeneracy regularization, which helps prevent representational collapse. Experiment settings are the same as \autoref{tab:eval_results}. }
\end{table}

\subsection{Ablations}
To help avoid degenerate solutions and ensure performant cluster assignments, Orchestra uses two important operations in its local training: degeneracy regularization and an EMA-based target model (see \S\ref{sec:method}). We now provide ablative results for these operations in \autoref{tab:ablations}. As can be seen, by removing either of the operations, Orchestra can lose substantial performance. As expected, noisy assignments can be overcome with time and hence Orchestra is less sensitive to the use of target model. On the other hand, degeneracy regularization is critical for Orchestra, as it helps prevent collapse of the model from the get-go.

\subsection{T-SNE Visualizations}
\begin{figure}
\centering
\begin{subfigure}{0.33\columnwidth}
  \centering
  \centerline{\includegraphics[width=\columnwidth]{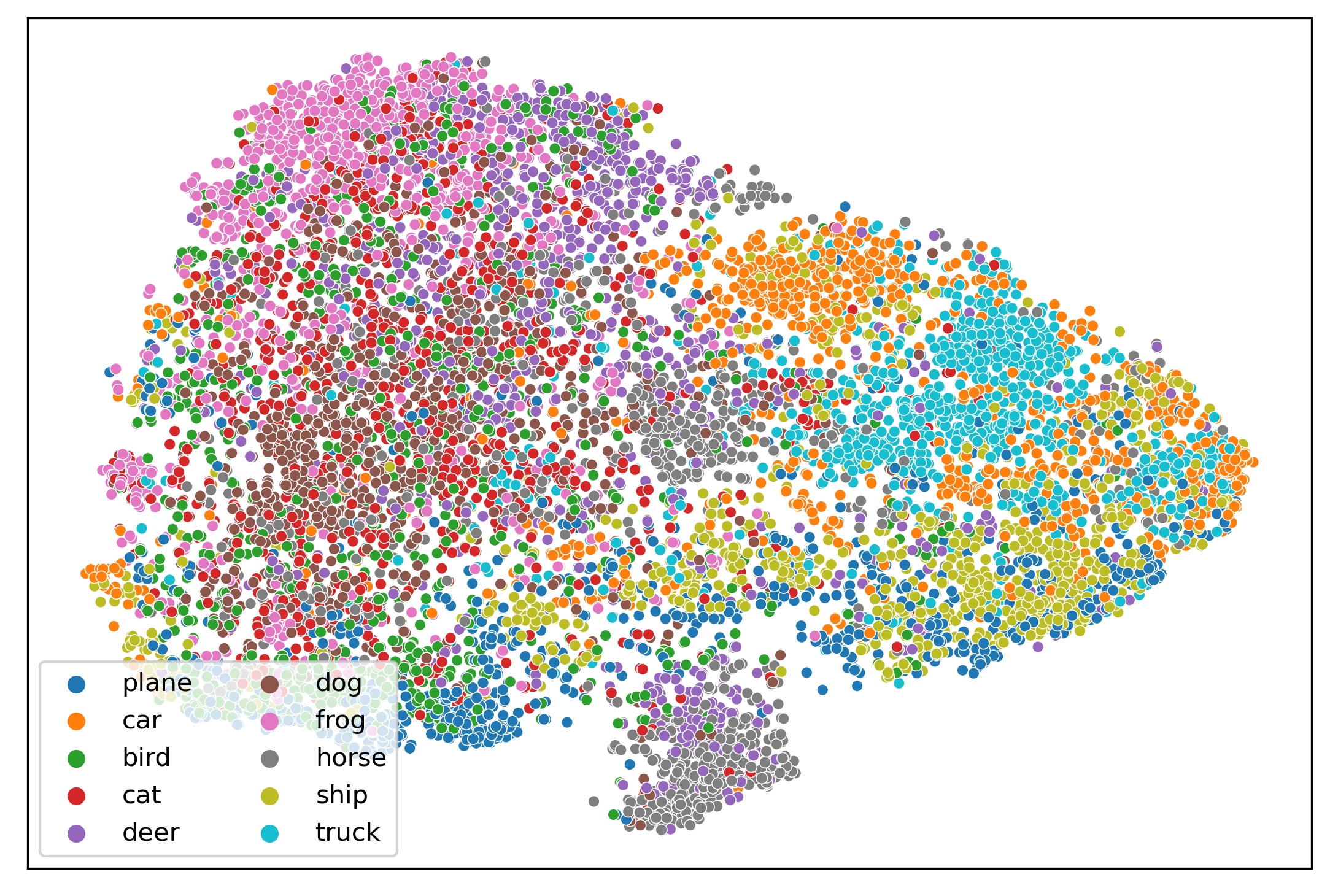}}
  \caption{f-BYOL}
\end{subfigure}%
\begin{subfigure}{0.33\columnwidth}
  \centering
  \centerline{\includegraphics[width=\columnwidth]{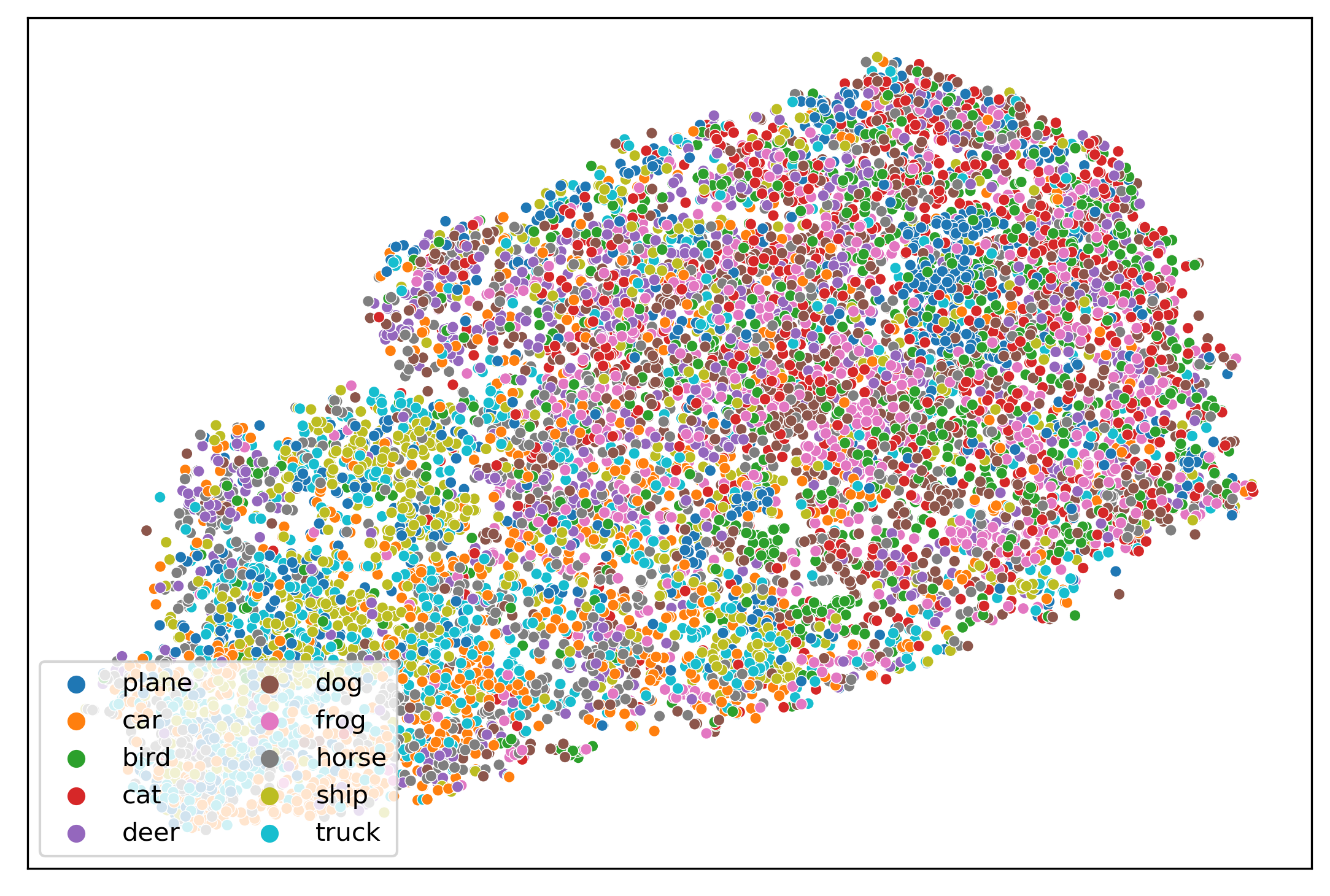}}
 \caption{RotPred}

\end{subfigure}
\begin{subfigure}{0.33\columnwidth}
  \centering
  \centerline{\includegraphics[width=\columnwidth]{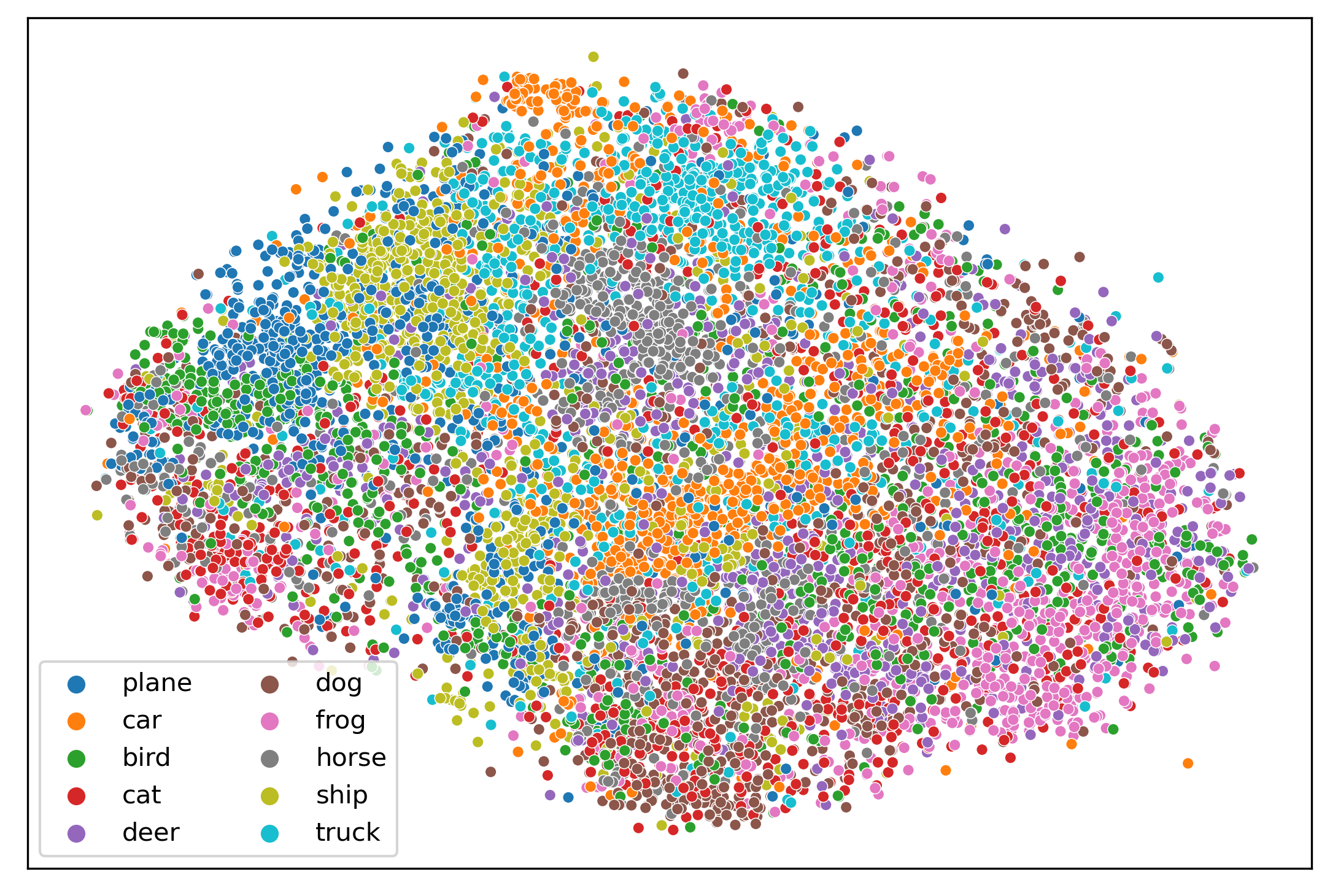}}
    \caption{f-SimCLR}

\end{subfigure}\\
\begin{subfigure}{0.33\columnwidth}
  \centering
  \centerline{\includegraphics[width=\columnwidth]{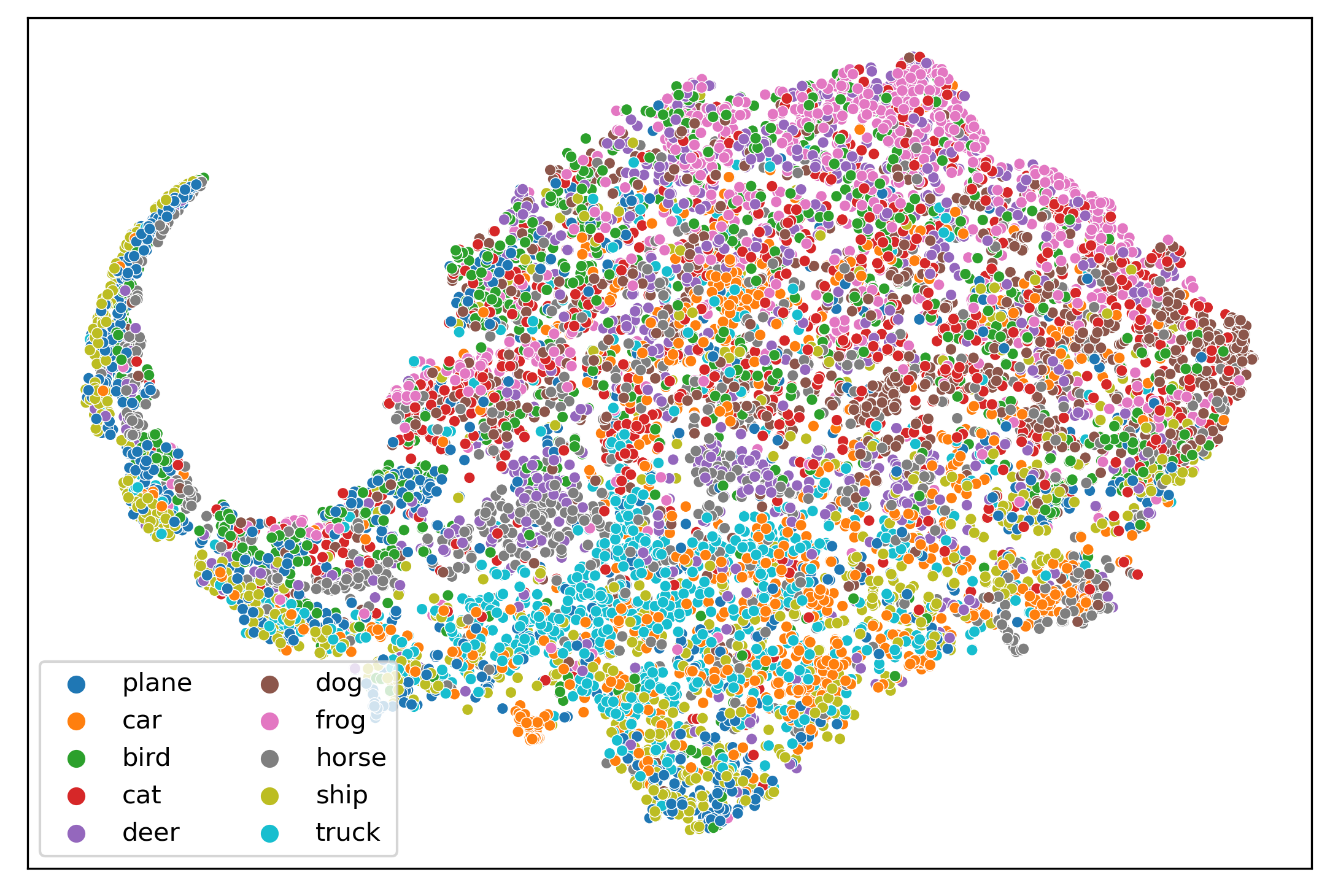}}
  \caption{f-SimSiam}

\end{subfigure}%
\begin{subfigure}{0.33\columnwidth}
  \centering
  \centerline{\includegraphics[width=\columnwidth]{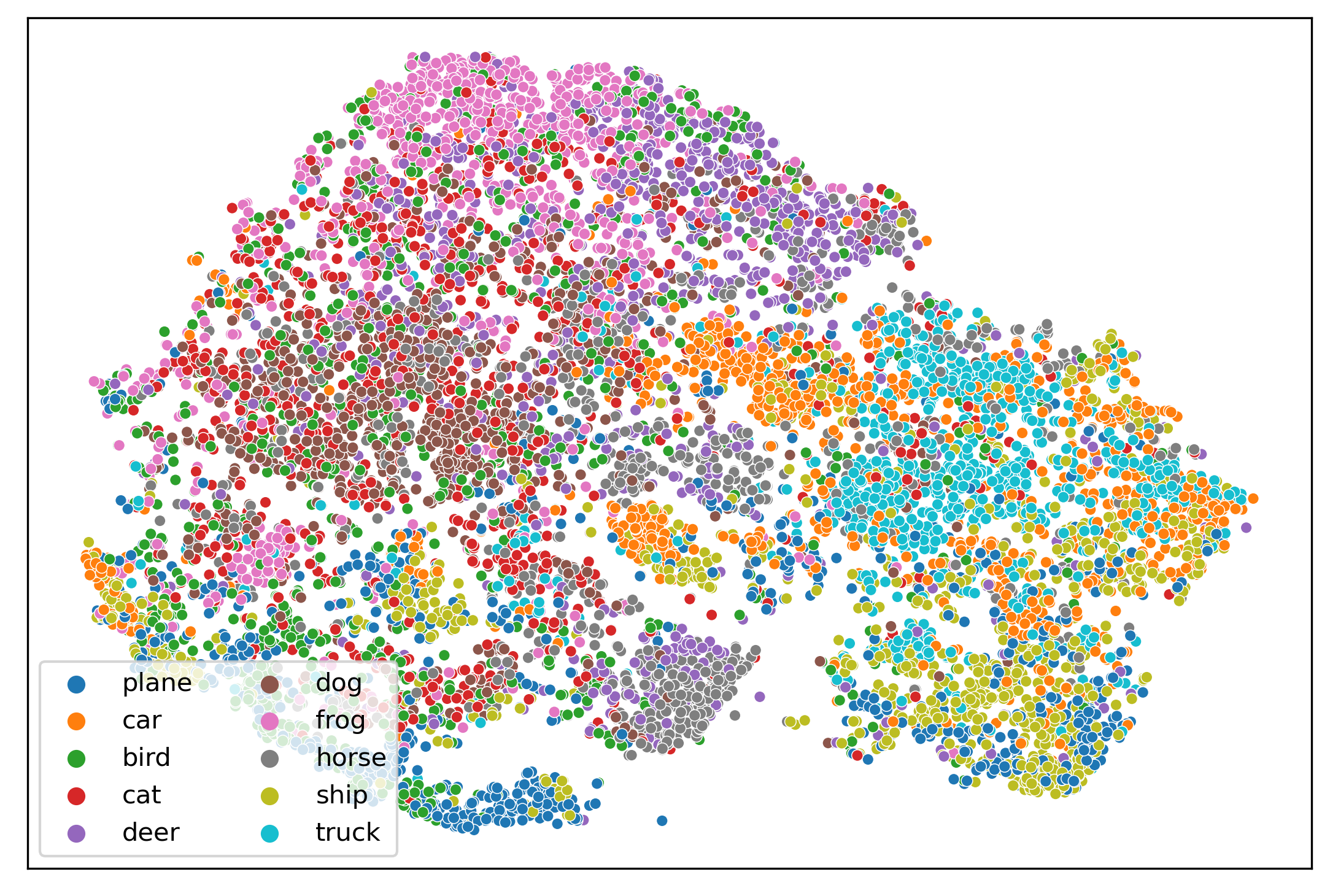}}
    \caption{f-SpecLoss}

\end{subfigure}
\begin{subfigure}{0.33\columnwidth}
  \centering
  \centerline{\includegraphics[width=\columnwidth]{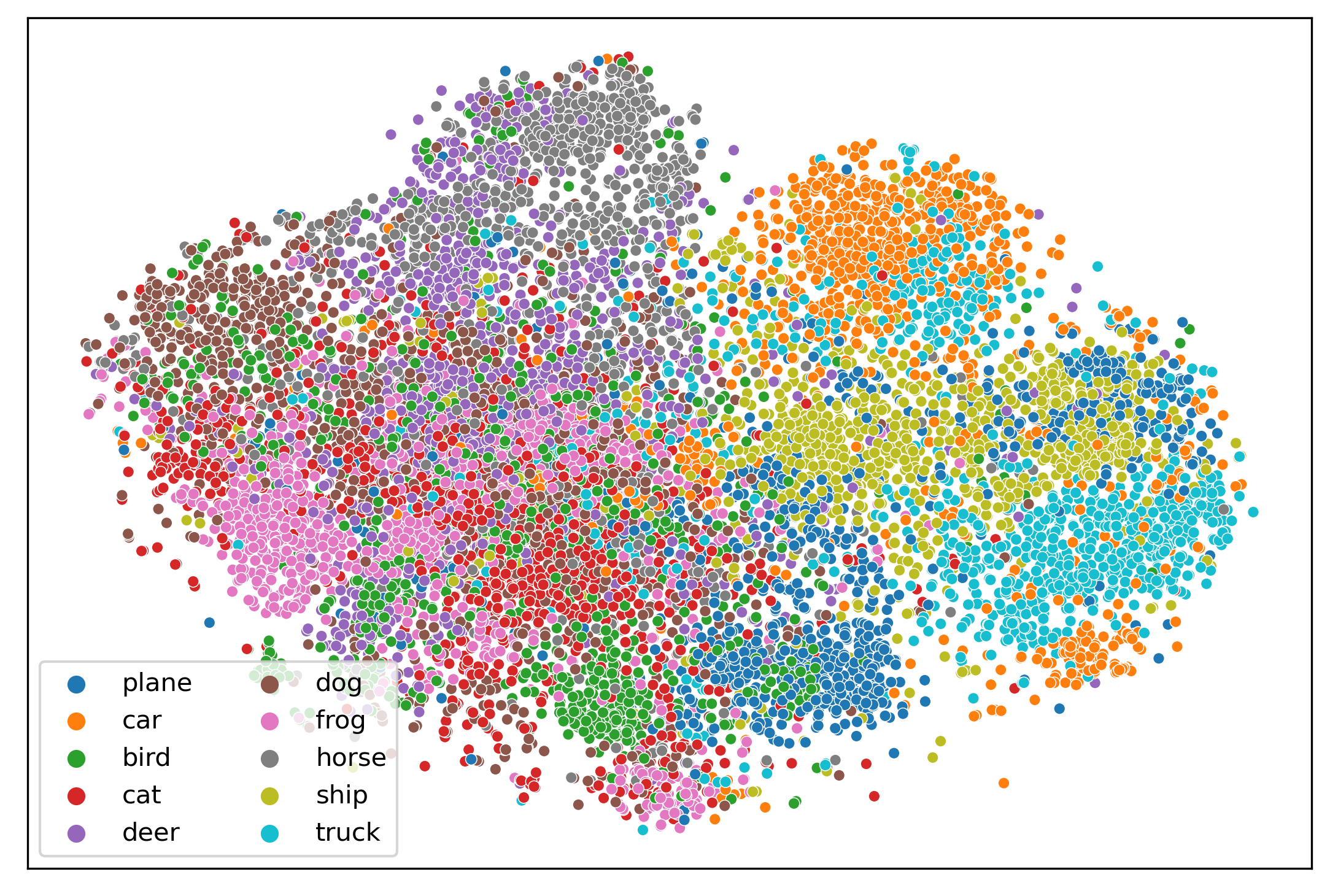}}
    \caption{\textbf{Orchestra}}

\end{subfigure} 
\vspace{-2mm}
\caption{TSNE visualizations obtained on the CIFAR-10 test set with models trained using different federated unsupervised learning approaches. No labeled data was used to fine-tune the models and the plot shows the quality of representations learned only using unlabeled data on the clients. }
\label{fig:tsne}
\vspace{-10pt}
\end{figure}

In Figure~\ref{fig:tsne}, we present the T-SNE visualizations obtained on the CIFAR-10 test set with models trained using different federated unsupervised learning approaches. These plots show the quality of representations learned only using unlabeled data on the clients. We can see that Orchestra provides better class separation than other methods.   


\subsection{Orchestra and Privacy}
\label{subsec:privacy}
Orchestra, by design, does not share any raw data or representations between clients and the server. Only the local centroids computed on each client are shared with the server for the purpose of global clustering. Below we discuss how the design of Orchestra aligns well with the idea of $K$-anonymous clustering, and how we can further improve Orchestra's privacy guarantees using local differentially private clustering. 

\textbf{K-anonymity:} We first focus on the $K$-anonymous clustering perspective. Formally, a $K$-anonymity guarantee ensures that for any randomly selected entry in a set, there are at least $K-1$ other entries with the same attributes. While in the discrete setting quasi-identifiers can be used to attack $K-$anonymity guarantees, these mechanisms provably fail in the continuous setting with high dimensional variables~\cite{kanonhardness}, making them particularly useful for our settings due to their good utility. $K$-anonymous clustering methods thus focus on finding clusters with at least $K$ members, providing non-uniform privacy to different samples. Our solution enforces an equal-size constraint on all $L$ local clusters using the sinkhorn-knopp based clustering algorithm~\cite{otclustering}. This enables uniform, $\nicefrac{N}{L}$-anonymity across all $N$ samples present on a client.   Further, as we showed in Table~\ref{tab:lg}, Orchestra is robust to the number of local clusters and can work with small values of $L$, which increases the anonymity guarantees of the algorithm. 

\textbf{Local Differential Privacy:} Differentially private (DP) algorithms~\cite{localdp} seek to design randomized mechanisms or algorithms with stochastic outputs by adding noise to the result of the mechanism. This guarantees that a given result could have been generated from multiple viable datasets. Formally, let $\mathcal{A}$ be a randomized mechanism that takes in a dataset $\mathcal{D}$ as input and whose image is denoted by the set $\mathcal{S}$. Assume $\mathcal{D}_1$ and $\mathcal{D}_2$ are two neighboring datasets, i.e., their entries differ in only one point. Then, $\mathcal{A}$ is ($\varepsilon, \delta$) differentially private if:
\begin{equation}
    \text{Pr}\left[ \mathcal{A}(\mathcal{D}_1) \in \mathcal{S} \right] \leq \exp(\varepsilon)\cdot \text{Pr}\left[ \mathcal{A}(\mathcal{D}_2) \in \mathcal{S} \right] + \delta
\end{equation}
In the situation a dataset is distributed across multiple participants, DP algorithms assume an honest server will conduct the algorithm by acquiring necessary information from the participants. This can be problematic in the situation where a server is only semi-honest and may seek to leak some sensitive information. To avoid this, \emph{local}-DP guarantees have been developed. In this case, one applies the DP definition for individual participants. For example, if ${x}_1$ and ${x}_2$ denote two participants, then local DP is defined as:
\begin{equation}
    \text{Pr}\left[ \mathcal{A}({x}_1) \in \mathcal{S} \right] \leq \exp(\varepsilon)\cdot \text{Pr}\left[ \mathcal{A}({x}_2) \in \mathcal{S} \right] + \delta
\end{equation}
Recently, local DP has been used for designing private clustering algorithms for distributed settings~\cite{dpbalcan, dpstemmer, dpchang}. These algorithms follow a standard route generally: (i) find a coreset that is representative of the dataset structure of a participant, but only approximately depends on it; (ii) use this coreset as an approximate notion of clusters from a participant; and (iii) find centroids that partition the coresets. The caveat with this approach is that since DP works on an aggregate scale, if a participant has few samples, the amount of noise that needs to be added to guarantee strong privacy can be huge, as shown by Cohen et al.~\yrcite{dpcohen}. Thus, even though one seeks to ensure $\varepsilon < 1.0$ and $\delta \sim \mathcal{O}\left(\frac{1}{\text{number of samples}}\right)$, production systems generally use $\varepsilon$ values of order $10.0$~\cite{census}. 

In \autoref{tab:dpprelims}, we provide linear probe results on CIFAR-10 for Orchestra with local-DP based clustering using the current state-of-the-art algorithm~\cite{dpchang}, which uses locality sensitive hashing for finding coresets in a participant's dataset. As can be seen, for practically useful values of $\varepsilon$, Orchestra witnesses only a small performance drop w.r.t.\ $K$-anonymity and is still outperforming alternative federated unsupervised methods (see \autoref{tab:eval_results}).

\begin{table}[h]
\centering
\begin{tabular}{@{}cccccc@{}}
\toprule
& No Privacy & $K$-anonymity & $\varepsilon$=0.8 & $\varepsilon$=1.0 & $\varepsilon$=10.0 \\ \midrule
Accuracy & 71.95 & 71.58 & 69.46 &69.60&69.63\\
\bottomrule
\end{tabular}
\caption{\label{tab:dpprelims}Accuracy obtained using the linear probe evaluation technique~\cite{simclr} in five different privacy settings. `No Privacy' represents the idealized setting when local representations are shared with the server -- this setting is prone to representation inversion attacks~\cite{invertcnn, invertautoregressive}. The K-anonymity result represents the performance of Orchestra with its 2-level clustering approach, however without any adding any DP protections. We can see that Orchestra's performance with its built-in K-anonymity is almost similar to the `No Privacy' setting. Further, Orchestra works reasonably well with local-DP based clustering; for practically useful values of $\varepsilon$, Orchestra witnesses only a small performance drop w.r.t. the K-anonymity setting, and still outperforms the various baseline techniques shown in  \autoref{tab:eval_results}. For these experiments, we use $\delta$=1e-3, Dirichlet $\alpha$=0.1, 100 clients, 10 local epochs, and 100 communication rounds.}
\end{table}

\section{Deferred Proofs}
We provide deferred proofs in this section. For better readability and reference, we first tabulate the notations used in the paper in Table~\ref{tab:notations}. 
\begin{table}[ht]
\centering
\begin{tabular}{lp{12cm}}
\toprule
Notation & Definition\\
\toprule
$X \sim \mathcal{X}$ & Unlabeled samples $X$ drawn from a distribution $\mathcal{X}$ \\ 
$K$ & Number of clients \\
$M$ & Number of classes \\
$N$ & Number of samples \\
$\mathcal{T}: \mathcal{X} \to \mathcal{\tilde{X}}$ & A stochastic augmentation function that transforms its input $x$ to the space of augmented samples by randomly selecting a transform from a set of predefined transformation functions that is finite, but can be large. \\
$\tilde{X} := \{\mathcal{T}(x): x \in X\}$ & The collective set of augmented samples \\
$f: \mathcal{X} \to \mathrm{R}^{D}$ & A parametric representation function \\
$\mathcal{E}(f)$ & Error of $f$ under a linear probe computed as $\min_{W} \mathbb{E}_{x\in\mathcal{X}}[\argmax(W^{T}f(x)) = y(x)]$\\ 
$R_{S} = \{f(s): s \in S\}$ & The set of representations on a set $S$ \\
$\mathcal{C}(\mathcal{B}, G)$ & A clustering algorithm that returns $G$ clusters on its input $\mathcal{B}$. \\
$\mu \in \mathrm{R}^{D \times G}$ & Centroids returned by the clustering algorithm $\mathcal{C}(\mathcal{B}, G)$.\\
$\mu_{g}$ & The centroid of cluster $g$ \\
$s_{f}(x)$ & Cosine similarity between centroids and representations $\nicefrac{\mu^{T}f(x)}{\norm{\mu^{T}f(x)}}$ \\
$P_{f}(x)$ & Cluster assignment probabilities computed as $\sigma(s_{f}(x))$ where $\sigma(.)$ denotes the softmax function\\ 
$\mathcal{H}(., .)$ & Cross-entropy between two discrete distributions \\
$\mathcal{L}_{\text{spec}}$ & Spectral Contrastive Loss~\cite{specloss}: $=-2\mathbb{E}_{x\in X, \tilde{x}\sim\mathcal{T}(x)}[s_f(x)^{T}s_f(\tilde{x})]+\mathbb{E}_{x, y\neq x\in X}[(s_f(x)^T s_f(y))^{2}]$ \\
$\mathcal{F}$ & Hypothesis class that has a global minimizer of the $L_{\text{spec}}$ \\
 \bottomrule
\end{tabular}
\caption{Notations used in the paper}
\label{tab:notations}
\end{table}

\subsection{Orchestra's pipeline reduces $\delta$}

In \S\ref{sec:method}, we claimed that by sharing the same set of centroids across all clients, Orchestra's pipeline reduces inter-cluster mixing $\delta$ every round. We formalize this result below.

\begin{proposition}
If the same set of global centroids $\mu$ are used across all clients, minimizing a loss that brings a sample's assigned centroids and its representation closer will ensure Orchestra's pipeline reduces $\delta$ every round.
\begin{proof}
It is easy to see Orchestra's pipeline is an Expectation-Maximization (EM) framework~\cite{embook}. Thus, a standard proof schematic for showing convergence of EM can be used. In particular, assume we compute set of $G$ global centroids from the local centroids $\mu = \mathcal{C}(\{\mathcal{C}(R_{X^{k}}, L^{(k)}): k \in [K]\}, G)$. Denote the set of samples assigned to global cluster $g$ as $\pi_{g}$. We overload the notation and let $\pi(x)$ denote the assignment of sample $x$. Then, without loss of generality, we let $(g^{t}_{m}, n^{t}) = \argmax_{g_m \in [G]}\argmax_{x_n \in \{X -  \pi_g\}} \mu_{g}^{T} f(x_{n})^{t}$, where the superscript denotes round $t$. That is, $\delta^t = \mu_{g_m^{t}}^{T}f(x_{n})^{t}$. During local training, if the similarity between $x_n$'s assigned centroid and its representation is increased, we have $\mu_{g_\pi^{t}(x_{n})}^{T}f(x_n)^{t+1} \geq \mu_{\pi^{t}(x_{n})}^{T}f(x_n)^{t}$, and, consequently, $\mu_{g_{m}^{t}}^{T}f(x_n)^{t+1} \leq \mu_{g^t_m}^{T}f(x_n)^{t} = \delta^t$ due to the use of softmax for computing assignments. With a similar argument for the server's clustering algorithm, we find the cluster centroids are brought closer to model representations that belong to that cluster. That is, $\mu_{g_{m}^{t+1}}^{T}f(x_n)^{t+1} \leq \mu_{g^t_m}^{T}f(x_n)^{t+1} \leq \delta^t$. Now, if $\delta^{t+1} = \mu_{g_m^{t+1}}^{T}f(x_{n})^{t+1}$, then we have $\delta^{t+1} \leq \delta^t$. If the sample and cluster for computing $\delta$ change, then without loss of generality, we assume $\delta^{t+1} = \mu_{g_{m'}^{t+1}}^{T}f(x_{n'})^{t+1} \leq \mu_{g_{m'}^{t}}^{T}f(x_{n'})^{t+1} \leq \mu_{g_{m'}^{t}}^{T}f(x_{n'})^{t} \leq \delta$. Here, the last inequality follows from definition of $\delta$. Second last inequality follows from the fact that local training brings representation of $f(x_{n'})$ closer to its assigned centroid and pushes it away from other centroids including $\mu_{g_{m'}}$. This is the step where we used the fact that global centroids are the same across all clients, due to which the above statement could be made on a dataset level. The first inequality follows from the global clustering step pushing $\mu_{g_{m'}}$ closer to representations of its assigned samples, which do not include $x_{n'}$.
\end{proof}
\end{proposition}
Note an assumption in the above argument is that all centroids are different from each other, else bringing a sample's representation closer to a centroid will not push it farther from the other centroids. In the case the model representations collapse to a single vector, as often observed in centralized clustering plus representation learning methods~\cite{towardskmeans}, all centroids become equivalent and hence our assumption is contradicted. Our use of a degeneracy regularization via predictive SSL was specifically motivated to enable this assumption, as it ensures representations do not collapse to a single vector. 

The above reasoning also explains why clustering-based SSL solutions \cite{swav, deepclusterv2, deepcluster, pcl} from centralized settings cannot be directly used in federated settings. Note that SwAV and related methods avoid degenerate solutions via \emph{per-iteration}, partition-based clustering. Consequently, using SwAV in FL would require communicating with the server every iteration. Since communication is very expensive in FL, SwAV becomes an infeasible baseline: for even 10 clients of CIFAR10, \emph{SwAV has 3100x higher communications costs than Orchestra's!} Upon using local training only (no per-iteration operations), we indeed found SwAV reaches degenerate solutions. 

\subsection{Proofs for Propositions~\ref{prop:distcluster} and \ref{prop:fedcluster}}
Our results are based on the analysis by HaoChen et al.~\yrcite{specloss}. Therein, the authors derive a general result that shows a minimizer of the loss $-2\mathbb{E}_{x\in X, \tilde{x}\sim\mathcal{T}(x)}[f(x)^{T}f(\tilde{x})]+\mathbb{E}_{x, y\in X}[(f(x)^T f(y))^{2}]$ will necessarily have small generalization error under a linear probe. This result arises out of an analysis of spectral clustering algorithms. As noted in the paper, our focus is partition-based clustering due to its straightforward application to distributed settings. Given the relationship between spectral clustering and partition-based clustering is well established~\cite{dhillon}, we can use the result by HaoChen et al.~\yrcite{specloss} for our purposes by showing Orchestra implicitly minimizes the objective above, consequently achieving small generalization error if it belongs to a sufficiently complex hypothesis class that can minimize SpecLoss. We first discuss the main concepts underlying those works.

\subsubsection{Background}
The works above include an important assumption called ``recoverability'', which is itself related to two important concepts called ``expansion'' and ``separation''. The notion of expansion and separation was originally proposed for manifolds by Balcan et al.~\yrcite{expansion}, recently verified empirically by Wei et al.~\yrcite{selftraining}, and converted to ``recoverability'' or graph connectivity properties by HaoChen et al.~\yrcite{specloss}. Specifically, the authors assume that there exists a latent graph instantiating the data distribution such that the neighborhood of any low probability set of connected nodes has a higher probability than the set itself. Here a node represents a sample and edges represent joint probability of occurrence of two samples. Under expansion, any given community of nodes on the latent graph is guaranteed to be sufficiently well connected such that if a node with high enough probability is assigned a label, then via a loss promoting consistency to input perturbations, the label will propagate throughout the community. If communities are assumed to be well separated, these labels will have higher diffusion within the same community than between different communities. This induces sets of abstract classes with semantically related nodes (e.g., community of dogs). Using standard pairwise clustering tools, one can compute node representations that enable discrimination between these communities and, if a sufficiently expressive parametric model (e.g., a neural network) is trained to match these representations, it can be guaranteed the model will have small generalization error on unseen data. The approach outlined above was recently used by HaoChen et al.~\yrcite{specloss} to design \emph{SpecLoss}, an SSL technique with provable generalization guarantees. However, since pairwise clustering requires computation of representation similarity between different datapoints, it is not amenable to use in decentralized, privacy-sensitive settings. In our work, we circumvented this issue by designing a partition-based, federated clustering framework that first computes centroids capable of partitioning the federation's data into discriminable clusters and then asks clients to allocate their data into these clusters.

\begin{assumption}
\label{asmp:recoverability} 
Recoverability. Let $\tilde{x} \in \mathcal{T}(x)$, where $x \sim \mathcal{X}$, and assume its ground truth label is $y(\tilde{x})$. We assume there exists a classifier $U$ such that $g(x) = y(\tilde{x})$ with probability at least $1-\phi$.
\end{assumption}

Let $\mathcal{G}$ denote a graph whose vertices $v_{x}$ represent samples from distribution $\mathcal{X}$ and whose edges denote joint probability of occurrence of the graph nodes, i.e., $w(x, \tilde{x}) = \mathbb{E}_{\tilde{x}_1, \tilde{x}_2 \in \mathcal{A}(x)}\left[\text{Pr}(\tilde{x}_1, \tilde{x}_2)\right]$
\begin{definition}
\label{defn:conductance} 
Sparsest $m$-partition. For an integer $m \in [2, |\mathcal{X}|]$, the sparsest $m$-partition is defined as $\rho_{m} := \min_{S_1, S_2, \dots, S_m} \max \{ \kappa_{\mathcal{G}}(S_1), \kappa_{\mathcal{G}}(S_2), \dots, \kappa_{\mathcal{G}}(S_m)\}$, where $\{S_1, S_2, \dots, S_m\}$ denotes non-empty sets that form a partition of $\mathcal{X}$ and $\kappa_{\mathcal{G}}(S_i):= \frac{\sum_{x \in S_i, x \not\in S_i} w(x x')}{\sum_{x \in S_i}w(x)}$ denotes the Dirichlet Conductance.
\end{definition}

Under these definitions, we have the following result.
\begin{theorem}
(Theorem 4.2 from HaoChen et al.~\yrcite{specloss}). Let Assumption~\ref{asmp:recoverability} hold, $f_{\text{pop}} \in \mathcal{F}$ is the population minimizer of $L_{\text{spec}}$, and assume $G\geq 4M+2$,. Define $\zeta_{\mathcal{X}} = \frac{\phi}{\rho_{\nicefrac{G}{2}}} \cdot \log(G)$. Then for an empirical minimizer function $f$ of SpecLoss such that $\mathcal{L}_{\text{spec}}(f) < \mathcal{L}_{\text{spec}}(f_{\text{pop}}) + \epsilon$, we have:
\begin{equation}
\label{eq:specbound}
    \mathcal{E}(f) < \zeta_{\mathcal{X}} + \mathcal{O}(\epsilon).
\end{equation}
\end{theorem}
Here, $\zeta_{\mathcal{X}}$ is a property of the data distribution $\mathcal{X}$ and $\mathcal{O}$ hides constants related to Rademacher complexity of the function class. Essentially, if the classes of augmentations of a sample can be predicted from the sample, then $\phi$ and hence $\zeta_{\mathcal{X}}$ are small. This happens if the latent variables instantiating the data generating distribution are similar so that the label of an augmented sample can be predicted from the original sample itself. Further, the denominator $\rho_{\nicefrac{G}{2}}$ depends on the \emph{average} probability of a partition. If we have $G < 2M$, $\rho_{\nicefrac{G}{2}}$ will be zero and hence the error can be arbitrarily large. If $G > 2M$, even though it \emph{can} get larger with $G$, it will essentially remain constant since one starts inducing subpartitions of abstract classes at this point.

The above theorem is our primary tool. Our idea is to show that instead of the pairwise clustering algorithm (spectral clustering) used by HaoChen et al.\yrcite{specloss}, one can use a partition-based clustering algorithm and exploit the result above to understand if good generalization is feasible in a more practical manner for federated settings. To this end, we will compute the loss achieved by a representation function that yields consistent representations over augmentations and can partition the set of representations into $G$ clusters with small inter-cluster mixing.

\subsubsection{Proposition \ref{prop:distcluster}}
\label{subsubsec:distcluster}
We now provide proof for Proposition~\ref{prop:distcluster}, restated below for convenience. 
\begin{proposition}
Assume $f\in \mathcal{F}$. Compute $\mathcal{G} > 4M+2$ clusters $\mu = \mathcal{C}(\{R_{X^{k}} : k \in [K]\}, G)$ s.t.\ all clusters are equally sized. Then, if $f$ minimizes $\mathcal{L} := \mathbb{E}_{k \in [K]}\left[\mathbb{E}_{x\in \mathcal{X}_{k},\tilde{x}\sim\mathcal{T}(x)}\left[\mathcal{H}\left(P_f(x), P_f(\tilde{x})\right)\right]\right]$, we have
\begin{equation}
\footnotesize
\mathcal{E}(f) < \zeta_{\mathcal{X}} + \mathcal{O} \left(2\delta + (G-2)\delta^2\right).
\end{equation}
\begin{proof}
Assume the hypothesis class is expressive enough to guarantee a zero error population minimizer exists for SpecLoss. Then, we need only bound the error of the empirical minimizer. We thus decompose $L_{\text{spec}} = L^{+} + L^{-}$, where we define the following terms $L^{+} := -\frac{2}{N |\mathcal{T}(x)|} \sum_{x \in X, \tilde{x} \in \mathcal{T}(x)} (s_f(x)^{T}s_f(\tilde{x}))$ and $L^{-} := \frac{1}{N(N-1)} \sum_{x \in X} \sum_{y \neq x, y \in X} (s_f(x)^{T}s_f(y))^{2}$. Let $\Delta_{i}$ denote a vector whose $i^{\text{th}}$ term is 1 and rest of the terms are $\delta$, the inter-cluster mixing. Then, we have the following:
\begin{equation}
\begin{split}
    L^{-} &= \frac{1}{N(N-1)} \sum_{x \in X} \sum_{y \neq x, y \in X} (s_f(x)^{T}s_f(y))^{2}\\
            &\leq \frac{1}{N(N-1)} \sum_{x \in X} \sum_{y \in X} (s_f(x)^{T}s_f(y))\\
            &= \frac{1}{N(N-1)} \left(\nicefrac{N}{G}\right)^{2} \sum_{x \in X} \sum_{y \in X} \left(\frac{s_f(x)^{T}s_f(y)}{\left(\nicefrac{N}{G}\right)^{2}}\right)\\
            &= \frac{N}{G^{2}(N-1)} \sum_{g_i} \sum_{g_j} \left(\sum_{x \in g_i}\frac{s_f(x)}{\left(\nicefrac{N}{G}\right)}\right)^{T}\left(\sum_{y \in g_j}\frac{s_f(y)}{\left(\nicefrac{N}{G}\right)}\right)\\
            &= \frac{N}{G^{2}(N-1)} \sum_{g_i} \sum_{g_j} \left(\mu^{T}\mu_{g_i}\right)^{T}\left(\mu^T\mu_{g_j}\right)\\
            &\leq \frac{N}{G^{2}(N-1)} \sum_{g_i} \sum_{g_j} \left(\Delta_{i}\right)^{T}\left(\Delta_{j}\right)\\
            &= \frac{N}{G(N-1)} \left[\left(1 + (G-1) \delta^2\right) + (G-1)\left(2 \delta + (G-2) \delta^2\right) \right]\\
            &= \frac{N}{(N-1)} \left[\left(\frac{1}{G}\right) + \left(1-\frac{1}{G}\right)\left(2 \delta + (G-1) \delta^2\right) \right]\\
            &= \mathcal{O}\left(2 \delta + (G-1) \delta^2\right)\\
\end{split}
\end{equation}
In the above, the first inequality follows from two facts: one, $s(.)$ is always unit norm, and hence the inner products are bound to be less than one, allowing us to ignore squares; two, self-interaction terms are positive, i.e., $s_f(x)^{T}s_f(x) > 0$. The second inequality follows from the definition of inter-cluster mixing. Note that $P_f(x) = P_f(\tilde{x})$ for $x, \tilde{x} \in \mathcal{T}(x)$ since $f$ is a minimizer of $\mathcal{L}$. Correspondingly, we have $s_{f}(x) = s_{f}(\tilde{x})$ since $s(.)$ is scale invariant and hence we have $L^{+} = 0$. Adding $L^{+}$ and $L^{-}$ provides us the value of $\epsilon$ which can be directly substituted into \autoref{eq:specbound} to complete the proof.
\end{proof}
\end{proposition}
Note that we hide two constants in the final expression: a constant additive factor $\frac{N}{G(N-1)}$ and a multiplicative factor $\frac{N}{N-1}\left(1 - \frac{1}{G}\right)$. The former will be close to 0 and the latter close to 1 for even moderately sized values of $N$ and $G$. 

\subsubsection{Proposition \ref{prop:fedcluster}}
We now provide proof for Proposition~\ref{prop:fedcluster}, restated below for convenience. 
\begin{proposition}
Assume $f\in \mathcal{F}$. Denote the set of local centroids as $\mu^{L} = \{\mathcal{C}(R_{X^{k}}, L^{(k)}): k \in [K]\}$ and compute new global centroids $\mu^{G} = \mathcal{C}(\mu^{L}, G)$ that are equally sized. Assume at least a fraction $c$ samples are ``consistently'' assigned, i.e., they match their assignments from the idealized setting. Then, if $f$ minimizes the loss $\mathcal{L} := \mathbb{E}_{k \in [K]}\left[\mathbb{E}_{x\in X_{k},\tilde{x}\sim\mathcal{T}(x)}\left[\mathcal{H}\left(P_f(x), P_f(\tilde{x})\right)\right]\right]$,
\begin{equation}
\footnotesize
\mathcal{E}(f) < \zeta_\mathcal{X} + \mathcal{O} \left(\gamma (1-c^{2}) + (2\delta + (G-1)\delta^2) \right).
\end{equation}
\begin{proof}
The proof follows essentially the same route as \autoref{subsubsec:distcluster}. The part of the argument that changes is the computation of $L^{-}$, where we now have to account for the inconsistent assignments $c' = 1-c$. First, we have the following $((\mu^G)^{T} f(x))^{T}((\mu^G)^{T} f(y)) \leq \Delta_{\text{same}} = 1 + (G-1)\delta^2 $ if $x$ and $y$ belong to the same global cluster in the idealized setting. Similarly, $((\mu^G)^{T} f(x))^{T}((\mu^G)^{T} f(y)) \leq \Delta_{\text{diff}} = 2\delta + (G-2)\delta^2 $. We also define $n_G = \nicefrac{N}{G}$ as the number of datapoints assigned to a cluster. This implies the number of samples leaving a cluster are at least $c' n_G$ and can consider equality for worst-case analysis. Define the term $D_{ij}$ as samples that were originally in global cluster $i$ in the idealized setting but have now been moved to global cluster $j$ due to the use of local clustering. Note that for any global cluster $g$, we have $\sum_{i} D_{gi} \leq c' n_G$, i.e., number of samples coming from different clusters must equal the number of samples that have left the given cluster. Again, for worst-case analysis, we can assume equality. Overall, we get the following.
\begin{equation}
\begin{split}
    L^{-} &\leq \frac{1}{N(N-1)} \left(\sum_{i \in [G]}\left(((1-c') n_G) (c'n_G) + G\sum_{j\neq i}D_{ij}\cdot (n_G - D_{ij}) \right) \Delta_{\text{same}}\right)\\ 
    &\quad\quad+ \frac{1}{N(N-1)} \left(\sum_{i \in [G]}\left(((1-c') n_G) (N - c'n_G - n_G) + \sum_{j \neq i} D_{ij}(N - 2*n_G + D_{ij}) \right) \Delta_{\text{diff}} \right)\\
        &= \frac{1}{N(N-1)} \left( \left(G (1-c') c' n_G^2 + G c' n_G^2 - (G-1)\left(\sum_{ij}D_{ij}^2\right) \right) \Delta_{\text{same}}\right)\\ 
    &\quad\quad+ \frac{1}{N(N-1)} \left( G (1-c') n_G (N - (1 + c')n_G) + G c' n_G (N - 2 n_G) + (G-1)\left(\sum_{ij}D_{ij}^2\right) \Delta_{\text{diff}} \right)\\
        &\leq \frac{1}{N(N-1)} \left( Gc'(2-c')n_G^2 \Delta_{\text{same}} + (NGn_G(1-c') - G n_G^2 (1 - c'^2) + c' G n_G (N - 2 c' n_G)) \Delta_{\text{diff}} \right).\\
\end{split}
\end{equation}
Substituting expressions for variable terms and simplifying, we get,
\begin{equation}
\begin{split}
        L^{-}  &\leq \frac{N}{(N-1)} \left(\frac{1}{G} \left(1 - \frac{G^2}{N^2}\right) + \frac{1}{G}c'(2 - c') + \left(\left(1-\frac{c'}{G}\right)^{2} - \frac{1}{G}\right) (2 \delta + (G - 1) \delta^2) \right)\\
            &= \frac{N}{(N-1)} \left(\frac{1}{G} \left(1 - \frac{G^2}{N^2}\right) + \frac{1}{G}(1 - c^2) + \left(\left(1-\frac{1-c}{G}\right)^{2} - \frac{1}{G}\right) (2 \delta + (G - 1) \delta^2) \right)\\
            &= \mathcal{O} \left(\gamma(1 - c^2) + (2 \delta + (G - 1) \delta^2) \right)\\
\end{split}
\end{equation}
Here, $\gamma = \frac{1}{(G - 1 -c)^2 - G}$ is a constant that is $<1$ for $G > 2$. Again adding and substituting $L^+$ and $L^-$ in \autoref{eq:specbound} finishes the proof.
\end{proof}
\end{proposition}
\end{document}